\documentclass{article}
\usepackage{amsmath}
\usepackage{amssymb}
\usepackage{xcolor}
\usepackage{colortbl}
\usepackage{booktabs}
\usepackage{caption}
\usepackage{subcaption}
\usepackage{float}
\usepackage{tikz}
\usepackage{tabularx} 
\usepackage{array} 
\PassOptionsToPackage{sort&compress}{natbib} 
\usepackage[preprint]{corl_2026} 

\makeatletter
\renewcommand{\paragraph}{%
  \@startsection{paragraph}{4}%
  {\z@}{0.50ex \@plus 1ex \@minus .2ex}{-1em}%
  {\normalfont\normalsize\bfseries}%
}
\makeatother

\title{What Are We Actually Benchmarking\\ in Robot Manipulation?}

\hypersetup{
  pdftitle={What Are We Actually Benchmarking in Robot Manipulation?},
  pdfauthor={Tianchong Jiang, Xiangshan Tan, Samuel Wheeler, Luzhe Sun, Tewodros W. Ayalew, Matthew Walter},
  pdfsubject={Robot manipulation benchmark audit preprint}
}

\author{
  \textbf{Tianchong Jiang}$^{1}$ \quad
  \textbf{Xiangshan Tan}$^{1,*}$ \quad
  \textbf{Samuel Wheeler}$^{3,*}$\\
  \textbf{Luzhe Sun}$^{1,*}$ \quad
  \textbf{Tewodros W.~Ayalew}$^{2,*}$ \quad
  \textbf{Matthew Walter}$^{1}$\\
  $^{1}$Toyota Technological Institute at Chicago, Chicago, IL, USA\\
  $^{2}$University of Chicago, Chicago, IL, USA\\
  $^{3}$Argonne National Laboratory, Lemont, IL, USA\\
  \texttt{\{tianchongj,luzhesun,vincenttann,mwalter\}@ttic.edu}\\
  \texttt{tewodrosayalew@uchicago.edu, swwheeler@anl.gov}\\
  $^{*}$Equal contribution.
}

\begin{document}

\maketitle


\begin{abstract}
A robotics benchmark score measures success under one fixed evaluation setup, yet is routinely treated as evidence of general manipulation capability.
We identify four failure modes, each of which weakens or invalidates a benchmark's role as a valid proxy for that capability: shortcut solvability, lack of statistical significance, creeping overfitting, and data-source dependence.
We propose one diagnostic per failure mode.
We audit LIBERO, CALVIN, SimplerEnv, RoboCasa, and RoboTwin~2.0 under these diagnostics.
LIBERO and CALVIN fail multiple diagnostics. RoboCasa and RoboTwin~2.0 fail fewer, despite appearing far less often in recent progress claims.
On LIBERO, a 0.09B probe with no language encoder scores at or near reported SOTA, and most reported gains are not provably statistically significant.
On CALVIN, randomizing block poses within the training range drops performance for every tested policy.
We release the four diagnostics with reference implementations for authors and reviewers to apply before treating a benchmark score as evidence of progress.
Code and artifacts are available at \href{https://ripl.github.io/manipulation_benchmark_audit/}{ripl.github.io/manipulation\_benchmark\_audit}.
\end{abstract}



\section{Introduction}\label{sec:introduction}

\begin{figure}[t]
    \centering
    \newcommand{\teaserTopHeight}{1.42in}
    \newcommand{\figOnePanelText}{\fontfamily{ptm}\fontsize{7pt}{8pt}\selectfont}
    \newcommand{\figOnePanelEmph}{\fontfamily{ptm}\fontsize{7pt}{8pt}\selectfont}
    \captionsetup{font={normalsize,rm}}
    \captionsetup[subfigure]{font={normalsize,rm}}
    \begin{subfigure}[t]{0.50\linewidth}
        \centering
        \includegraphics[height=\teaserTopHeight]{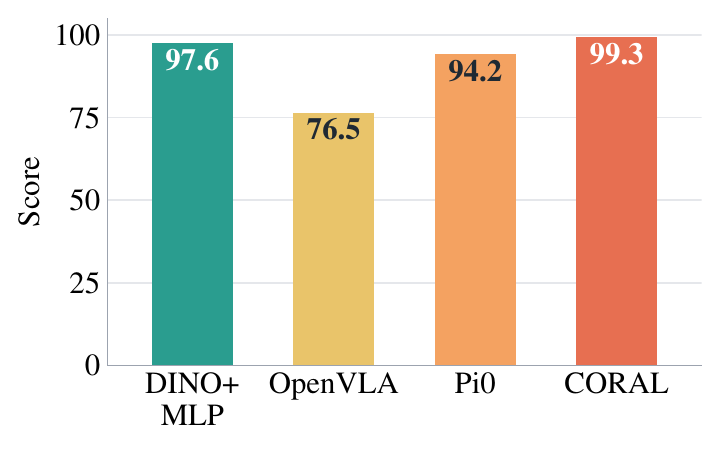}
        \caption{Shortcut solvability}
        \label{fig:teaser-shortcut}
    \end{subfigure}
    \hfill
    \begin{subfigure}[t]{0.48\linewidth}
        \centering
        \includegraphics[height=\teaserTopHeight]{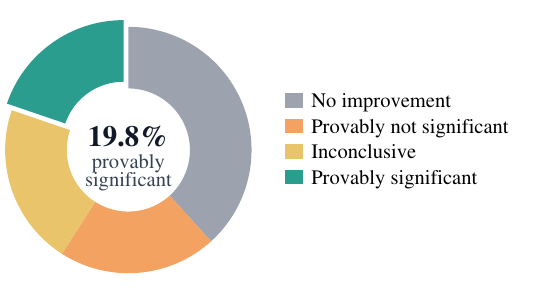}
        \caption{Statistical significance}
        \label{fig:teaser-significance}
    \end{subfigure}
    \par\vspace{-0.045in}
    \begin{subfigure}[t]{0.48\linewidth}
        \centering
        \begin{tikzpicture}[scale=0.81,font=\figOnePanelText]
        \definecolor{cXVLA}{HTML}{E76F51}
        \definecolor{cGR1}{HTML}{F4A261}
        \definecolor{cRF}{HTML}{E9C46A}
        \definecolor{cPieNoImprovement}{HTML}{9CA3AF}
        \colorlet{cAxis}{cPieNoImprovement}
        \colorlet{cSoftText}{cPieNoImprovement}
        \begin{scope}
            \def\ys{0.64}\def\xa{1.00}\def\xb{2.45}
            \draw[cAxis,semithick,->] (0,0) -- (0,3.10);
            \draw[cAxis,semithick,->] (0,0) -- (2.85,0);
            \node[rotate=90,anchor=south,cSoftText] at (-0.38,2*\ys) {Avg. tasks completed (ATC)};
            \foreach \v in {0,1,2,3,4} {
            \draw[cAxis] (-0.05,\v*\ys) -- (0,\v*\ys);
            \node[left,cSoftText] at (-0.07,\v*\ys) {\v};
            }
            \draw[cAxis] (\xa,0) -- (\xa,-0.05);
            \draw[cAxis] (\xb,0) -- (\xb,-0.05);
            \node[below,cSoftText] at (\xa,-0.07) {official};
            \node[below,cSoftText] at (\xb,-0.07) {resampled};
            \draw[cXVLA,line width=1.3pt] (\xa,4.165*\ys) -- (\xb,3.138*\ys);
            \fill[cXVLA] (\xa,4.165*\ys) circle (2.0pt); \fill[cXVLA] (\xb,3.138*\ys) circle (2.0pt);
            \node[cXVLA,anchor=east] at (\xa-0.08,4.165*\ys) {4.17};
            \node[cXVLA,anchor=west] at (\xb+0.04,3.138*\ys) {3.14};
            \node[cXVLA,anchor=south,font=\figOnePanelEmph] at (\xb,3.138*\ys+0.18) {X-VLA};
            \draw[cGR1,line width=1.3pt] (\xa,3.244*\ys) -- (\xb,2.495*\ys);
            \fill[cGR1] (\xa,3.244*\ys) circle (2.0pt); \fill[cGR1] (\xb,2.495*\ys) circle (2.0pt);
            \node[cGR1,anchor=east] at (\xa-0.08,3.244*\ys) {3.24};
            \node[cGR1,anchor=west] at (\xb+0.04,2.495*\ys) {2.50};
            \node[cGR1,anchor=west,font=\figOnePanelEmph] at (\xb+0.72,2.495*\ys) {GR-1};
            \draw[cRF,line width=1.3pt] (\xa,2.367*\ys) -- (\xb,1.869*\ys);
            \fill[cRF] (\xa,2.367*\ys) circle (2.0pt); \fill[cRF] (\xb,1.869*\ys) circle (2.0pt);
            \node[cRF,anchor=east] at (\xa-0.08,2.367*\ys) {2.37};
            \node[cRF,anchor=west] at (\xb+0.04,1.869*\ys) {1.87};
            \node[cRF,anchor=north,font=\figOnePanelEmph] at (\xb,1.869*\ys-0.09) {RoboFlamingo};
        \end{scope}
        \begin{scope}[xshift=3.75cm,yshift=0.00cm]
            \draw[cAxis,semithick,fill=gray!4] (0,1.79) rectangle (3.20,3.10);
            \fill[gray!60] (1.782,2.008) circle (0.045);
            \fill[gray!60] (2.436,2.008) circle (0.045);
            \fill[gray!60] (0.724,2.781) circle (0.045);
            \fill[gray!60] (1.856,2.781) circle (0.045);
            \node[anchor=north east,cSoftText] at (3.08,2.96) {official};
            \draw[cAxis,semithick,fill=gray!4] (0,0) rectangle (3.20,1.31);
            \foreach \x/\y in {2.380/0.339,1.993/0.520,1.750/0.350,2.419/0.219,1.803/0.277,2.521/0.262,2.386/0.505,2.519/0.174,2.341/0.274,1.706/0.313,2.510/0.339,2.328/0.143,2.811/0.114,2.107/0.513,2.852/0.205,2.411/0.404,2.182/0.331,2.403/0.116,1.750/0.293,2.766/0.149,2.260/0.371,2.538/0.202,1.818/0.325,1.703/0.368,2.856/0.397,2.210/0.349,1.855/0.377,2.361/0.462,2.835/0.242,2.809/0.172,2.472/0.393,2.417/0.133,2.426/0.194,2.142/0.347,2.155/0.177,2.294/0.176,2.778/0.422,2.777/0.517,2.621/0.307,1.892/0.181,1.643/0.421,2.502/0.248,2.313/0.521,2.368/0.152,2.359/0.258,2.757/0.238,2.450/0.321,2.314/0.362,2.211/0.211,2.151/0.387,2.095/0.180,2.303/0.369,1.686/0.139,2.833/0.355,1.827/0.255,2.341/0.179,2.215/0.353,1.663/0.420,2.386/0.489,1.681/0.450,2.771/0.345,2.691/0.124,2.061/0.291,2.130/0.268,2.718/0.136,1.765/0.483,2.113/0.350,1.763/0.483,2.400/0.403,2.730/0.415,2.457/0.248,2.863/0.252,1.703/0.491,2.320/0.291,2.098/0.498,1.736/0.411,1.979/0.378,1.772/0.389,2.840/0.529,2.470/0.122} { \fill[gray!60] (\x,\y) circle (0.045); }
            \fill[gray!60] (0.724,0.991) circle (0.045);
            \fill[gray!60] (1.856,0.991) circle (0.045);
            \node[anchor=north east,cSoftText] at (3.08,1.09) {resampled};
        \end{scope}
        \end{tikzpicture}
        \caption{Creeping overfitting}
        \label{fig:teaser-overfitting}
    \end{subfigure}
    \hfill
    \begin{subfigure}[t]{0.48\linewidth}
        \centering
        \includegraphics[height=\teaserTopHeight]{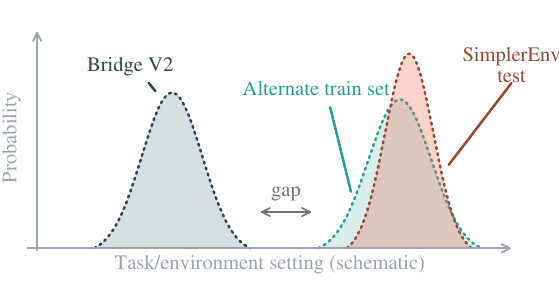}
        \caption{Data-source dependence}
        \label{fig:teaser-data-source}
    \end{subfigure}

    \caption{
    The panels preview the four diagnostics used in our audit.
    \subref{fig:teaser-shortcut}~A 0.09B-parameter probe nearly matches SOTA on LIBERO average.
    \subref{fig:teaser-significance}~Only 19.8\% of LIBERO SOTA claims are provably significant.
    \subref{fig:teaser-overfitting}~On CALVIN, resampling block poses within the training distribution lowers average tasks completed (ATC).
    \subref{fig:teaser-data-source}~On SimplerEnv, the score cannot distinguish a policy that generalizes from one trained on data close to the test.
    }
    \label{fig:diagnostic-teaser}
\end{figure}

In March 2026 alone, 79 arXiv papers reported results on LIBERO (Figure~\ref{fig:benchmark-result-paper-counts}).
When a benchmark is used at this rate, it is not only a testbed for individual researchers but also the field's definition of progress~\cite{Torralba2011UnbiasedLook}.

A benchmark score is not an end in itself.
It is a proxy for general capabilities that we want robots to have in the real world, like tidying our homes, cleaning and folding laundry, and assisting the elderly.
A benchmark is meaningful when a high score is evidence of that capability.
It is therefore important to test whether the most-reported manipulation benchmarks are meaningful.

The ideal test would train policies on shared demonstrations, evaluate them on a set of representative real-world tasks, and check whether the relative ranking of policies on the benchmark matches that on real-world tasks.
Such a test is impractical: there is no agreed-upon real-world task set to evaluate on, and retraining each policy on shared demonstrations is too expensive.
Another approach is to check whether policies that rank highly on one benchmark also rank highly on others~\cite{Torralba2011UnbiasedLook}.
This does not work either: Authors choose which benchmarks to report and tune their methods to reach SOTA on those, so new papers almost always claim SOTA on every benchmark they report.

We instead define four diagnostics that provide insight into the extent to which performance on a benchmark is evidence of general manipulation capability. We apply them to the five most widely reported manipulation benchmarks, LIBERO~\cite{Liu2023LIBERO}, CALVIN~\cite{Mees2022CALVIN}, SimplerEnv~\cite{Li2025SimplerEnv}, RoboCasa~\cite{Nasiriany2024RoboCasa}, and RoboTwin~2.0~\cite{Chen2025RoboTwin2}.

\noindent\textbf{Shortcut solvability.}
A benchmark score is evidence of capability only if any policy achieving it has the capability.
We show that for LIBERO, a simple 0.09B probe with no language encoder and no large-scale robotics pretraining scores at or near the best reported result, while on RoboTwin~2.0 and RoboCasa it scores well below the best reported result.

\noindent\textbf{Statistical significance.}
Most benchmarks report only an aggregate success rate, from which the significance of a reported gain cannot be tested.
On LIBERO and SimplerEnv, only 19.8\% and 19.7\% of SOTA claims, respectively, are provably significant.
On RoboCasa and RoboTwin~2.0, the shares rise to 53.3\% and 73.7\%, respectively.

\noindent\textbf{Creeping overfitting.}
A benchmark's test distribution can be a narrow region of the training distribution's support, and its fixed test set is one sample from that region.
Policies can fit either, which we call \emph{distribution overfitting} and \emph{sample overfitting}.
On CALVIN, resampling block poses within the training range lowers average tasks completed out of five (ATC) from 4.17 to 3.14. 

\noindent\textbf{Data-source dependence.}
A small model trained on demonstrations closer to the test distribution can match a much larger model trained on more general data.
As a result, a benchmark score cannot distinguish generalization from how close the training data is to the test distribution.

Across these four diagnostics, the most-reported benchmarks fail the most. LIBERO, CALVIN, and SimplerEnv each fail several diagnostics, while RoboTwin~2.0 and RoboCasa fare better.

We release reference implementations of the four diagnostics.
Authors of new benchmarks can run them before publishing.
For the five benchmarks we audit, our per-benchmark results help authors and reviewers interpret reported gains.
Code and artifacts are available through the project website: \href{https://ripl.github.io/manipulation_benchmark_audit/}{ripl.github.io/manipulation\_benchmark\_audit}.


\section{Related Work}
\label{sec:related-work}

\subsection{Benchmark Audits in Computer Vision and NLP}
\label{sec:cv-nlp-audits}

There is a rich literature auditing benchmark validity in computer vision and natural language processing, and each of the four failure modes we apply to manipulation has direct precedent there.

\noindent\textbf{Shortcut solvability.}
\citet{Goyal2017VQA2} showed that the original visual question answering (VQA) benchmark allowed models to answer questions starting ``Do you see a \ldots'' with 87\% accuracy by blindly answering ``yes'', without reading the rest of the question or looking at the image.
They built the VQA~v2 benchmark with complementary-image pairs
to remove this language-prior shortcut.
\citet{Lapuschkin2019CleverHans} found that models trained on PASCAL VOC could predict ``horse'' from a copyright tag in the corner of the image.
Analogously, on LIBERO and CALVIN, a probe with no language encoder can do relatively well.

\noindent\textbf{Statistical significance.}
\citet{Everingham2010VOC} applied the Friedman/Nemenyi test to PASCAL VOC and found no statistically significant difference among the top six competing methods.
\citet{Musgrave2020MetricLearningReality} later showed that a subfield's reported gains did not survive fair-comparison auditing.

\noindent\textbf{Creeping overfitting.}
Creeping overfitting is when models become too adapted to a benchmark over time, memorizing its idiosyncrasies and failing to generalize~\cite{Torralba2011UnbiasedLook,Ponce2006DatasetIssues}.
\citet{Torralba2011UnbiasedLook} showed this across popular vision datasets, where classifiers trained on one dataset transferred poorly to others, with average drops in accuracy near 50\%.
\citet{Recht2019ImageNetGeneralize} asked whether models had fit the one fixed test set.
They rebuilt a fresh test set under the original ImageNet protocol and found rankings largely unchanged, so little sign of fitting to that set.
\citet{Hendrycks2019ImageNetC} asked whether models had instead fit the narrow test distribution.
They corrupted the inputs and found accuracy dropped sharply, a clear sign of fitting to that distribution.
We run both checks on manipulation benchmarks.

\noindent\textbf{Data-source dependence.}
Several audits show that test-set scores partly reflect train-test proximity rather than learned capability.
\citet{Barz2020CIFAR} found that CIFAR-10 and CIFAR-100 test sets contain near-duplicates of training images, so reported accuracy partly measures memorization rather than generalization.
\citet{Elangovan2021Leakage} document analogous train/test leakage across NLP benchmarks.
\citet{Lewis2021QAOverlap} show that most test questions in popular open-domain QA datasets have an answer or near-paraphrase in the training set, and models score much worse on the leakage-free subset.
We extend this to robot manipulation.
A small model trained on data close to the test distribution can outperform a larger model trained on more distant data.

\subsection{Manipulation Benchmarks and Prior Audits}
\label{sec:manipulation-benchmarks}

LIBERO~\cite{Liu2023LIBERO} is a language-conditioned manipulation benchmark with four task suites.
CALVIN~\cite{Mees2022CALVIN} is a long-horizon language-conditioned manipulation benchmark with held-out instructions.
SimplerEnv~\cite{Li2025SimplerEnv} reproduces real Bridge and Google-robot setups in simulation.
RoboCasa~\cite{Nasiriany2024RoboCasa} is a kitchen-task benchmark with procedurally generated scenes and objects.
RoboTwin~2.0~\cite{Chen2025RoboTwin2} is a dual-arm manipulation benchmark with strong domain randomization for bimanual manipulation.
\begin{figure}[!t]
    \centering
    \includegraphics[width=0.98\linewidth]{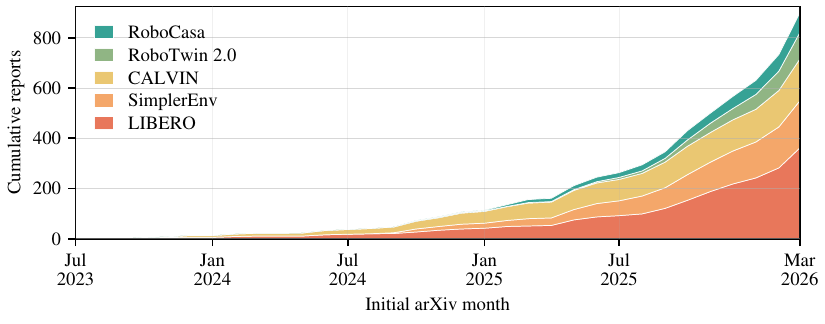}
    \caption{
        Cumulative counts of arXiv papers reporting results on CALVIN, LIBERO, SimplerEnv, RoboCasa, and RoboTwin~2.0.
        Sources, deduplication keys, and per-paper classification criteria are in Appendix~\ref{app:tracker-methodology}.
    }
    \label{fig:benchmark-result-paper-counts}
\end{figure}

Several prior works audit specific manipulation benchmarks.
LIBERO-Pro~\cite{Zhou2025LIBEROPRO} and LIBERO-Plus~\cite{Fei2025LIBEROPlus} change LIBERO scenes and instructions and show that high scores drop.
But their changed inputs fall outside the benchmark's training distribution, so a drop might be the result of weak generalization.
Our diagnostic instead keeps every input inside the training distribution, so a drop measures creeping overfitting rather than weak generalization.

Many other manipulation simulators~\cite{Zhu2020Robosuite,Mu2021ManiSkill,Shen2021iGibson} and benchmarks~\cite{Yu2020MetaWorld,Zeng2020Ravens,Jiang2023VIMA,Lynch2023LanguageTable,Gong2023ARNOLD,Zhang2024VLABench,James2020RLBench,Pumacay2024Colosseum,Heo2023FurnitureBench,Shridhar2020ALFRED,Li2023Behavior1K,Szot2021Habitat} exist.
We focus on the five in Figure~\ref{fig:benchmark-result-paper-counts} since they appear far more often in recent VLA and generalist-policy work (Appendix~\ref{app:tracker-methodology}).


\section{Shortcut Solvability}
\label{sec:shortcut-solvability}

A benchmark score is evidence of capability only if every policy that reaches a comparable score has the capability.
We call a policy that reaches the score while lacking the capability a \emph{shortcut}, and a benchmark \emph{shortcut-solvable} when one exists.

LIBERO~\cite{Liu2023LIBERO} is a language-conditioned manipulation benchmark that tests lifelong learning across language-specified tasks.
Scoring well is meant to require understanding the instruction, yet a high score is read as evidence of much more.
Because the top of its leaderboard is held by foundation VLAs, a high score is taken to show that the policy generalizes broadly, and that reaching it requires large-scale robotics pretraining and a large, expressive model.

We test whether a high score can be reached without those properties.
Our probe is a deliberately small model with approximately 90M parameters (a shared DINOv2 encoder~\cite{Oquab2023DINOv2} and a small MLP head), with no language encoder, no large-scale robotics pretraining, and no expressive action head (i.e., no diffusion head or action tokenizer).
Because each benchmark draws its instructions from a fixed public set, the probe replaces the instruction with a learned per-task embedding that it looks up by index instead of reading as text. On LIBERO each instruction maps to its own embedding, and on CALVIN several phrasings of a subtask map to the same embedding. 
The probe uses only what each benchmark's evaluation protocol lets any policy use, so reaching its score this way reveals a weakness in the protocol, not a trick of ours.

\begin{table}[!htb]
    \centering
    \captionsetup{font=footnotesize}
    \caption{Shortcut-solvability probe results across five benchmarks.}
    \label{tab:benchmark-audit-matrix}
    \scriptsize
    \setlength{\tabcolsep}{4pt}
    \renewcommand{\arraystretch}{1.12}
    \newcommand{\posdelta}[1]{\textcolor{green!35!black}{($#1$)}}
    \newcommand{\negdelta}[1]{\textcolor{red!45!black}{($#1$)}}
    \newcommand{\zerodelta}[1]{\textcolor{black!65}{($#1$)}}
    \newcommand{\benchrule}{\arrayrulecolor{black!30}\midrule\arrayrulecolor{black}}
    \begin{tabular}{@{}llrlrlr@{}}
    \toprule
    Benchmark & Protocol/Suite & DINO+MLP & Foundation VLA & Reported ($\Delta_F$) & Best Reported & Reported ($\Delta_B$) \\
    \midrule
    LIBERO & Spatial & 99.0\% & OpenVLA (7B) & 84.7\% \negdelta{-14.3} & $\pi_0$+T-MEE (2.3B) & 99.8\% \posdelta{+0.8} \\
           & Object & 100.0\% & OpenVLA (7B) & 88.4\% \negdelta{-11.6} & $\pi_{0.5}$+MoH (3B) & 100.0\% \zerodelta{+0.0} \\
           & Goal & 98.8\% & OpenVLA (7B) & 79.2\% \negdelta{-19.6} & MoLA (1.64B) & 99.5\% \posdelta{+0.7} \\
           & Long & 92.4\% & OpenVLA (7B) & 53.7\% \negdelta{-38.7} & CORAL$_{\text{SimVLA}}$ (0.8B) & 98.8\% \posdelta{+6.4} \\
    \benchrule
    CALVIN$^\dagger$ & D$\to$D & 3.123 & DeeR-VLA (3B) & 2.83 \negdelta{-0.29} & MDT-V (n.d.) & 4.52 \posdelta{+1.40} \\
                     & ABC$\to$D & 3.242 & RoboFlamingo (3B) & 2.48 \negdelta{-0.76} & MMaDA-VLA (8B) & 4.78 \posdelta{+1.54} \\
                     & ABCD$\to$D & 3.872 & RoboFlamingo (3B) & 4.09 \posdelta{+0.22} & Xiaomi-0 (4.7B) & 4.80 \posdelta{+0.93} \\
    \benchrule
    SimplerEnv & WidowX & 0.0\% & X-VLA (0.9B) & 95.8\% \posdelta{+95.8} & CORAL$_{\text{SimVLA}}$ (0.8B) & 97.9\% \posdelta{+97.9} \\
    \benchrule
    RoboTwin~2.0 & Clean (50+500) & 60.4\% & X-VLA (0.9B) & 72.8\% \posdelta{+12.4} & MotuBrain (n.d.) & 95.8\% \posdelta{+35.4} \\
                & Randomized (50+500) & 59.4\% & X-VLA (0.9B) & 72.8\% \posdelta{+13.4} & MotuBrain (n.d.) & 96.1\% \posdelta{+36.7} \\
    \benchrule
    RoboCasa & RSS24 & 18.8\% & GR00T-N1 (2B) & 49.6\% \posdelta{+30.8} & X-WAM (n.d.) & 79.2\% \posdelta{+60.4} \\
    \bottomrule
    \end{tabular}
    \par\vspace{0.3em}
    \begin{minipage}{0.95\linewidth}
    {\scriptsize \textbf{Foundation VLA}: most-cited protocol-comparable foundation VLA reporting on the protocol (chosen per protocol for CALVIN).\\
    \textbf{Best reported}: top non-RL-finetuned result (RL-finetuned excluded, as the probe is imitation-only).\\
    $\Delta_F$ and $\Delta_B$: the Foundation-VLA score and the Best-reported score, respectively, minus the DINO+MLP probe score (ATC for CALVIN, percentage points otherwise); a negative value means the probe scores higher.\\
    CALVIN$^\dagger$ reports ATC (out of 5). ``n.d.''~=~params not disclosed. Policy sources are listed in Appendix~\ref{app:experiment-details-shortcut}.}
    \end{minipage}
\end{table}
Table~\ref{tab:benchmark-audit-matrix} shows that LIBERO is shortcut-solvable since the probe performs within one percentage point of the best published result on three of four splits.\footnote{We train and tune the probe per suite and report the best of several checkpoints evaluated on the suite. See Appendix~\ref{app:experiment-details-shortcut}.}
On CALVIN, the shortcut is real but smaller, since the probe matches the performance of  widely-used policies but not the current best.
On RoboTwin~2.0, RoboCasa, and SimplerEnv, the probe scores well below the best, indicating the lack of a shortcut.

We claim only that a high score on LIBERO is not, on its own, evidence of general manipulation capability---our probe achieves a high score yet lacks such a capability.
We do not claim that a high-scoring policy lacks that capability, since a capable policy reaches a high score too.
\footnote{Suppose that one can pass a written driving test by memorizing the answer key. Passing this test is not evidence of driving ability. However, the people who pass the test can still be good drivers, just like the models that do well on LIBERO can still be capable.}


\section{Statistical Significance}
\label{sec:statistical-significance}

When a new policy is published, it will often report a higher aggregate benchmark score than the previous best from the literature, and that gap is taken as progress. In this section, we consider the question of determining  how statistically significant such performance gaps may be.  

Most papers report only an aggregate score, most commonly an average success rate. However, the aggregate score alone is often not enough to determine the statistical significance of a reported performance gap, a gap we then call \emph{inconclusive}. It arises because the variance of the empirically observed performance gap is a function of task-level and sample-level information that is not retained by the average performance statistic (Appendix~\ref{app:bounded-score-wald-test}).

\begin{figure}[H]
    \centering
    \includegraphics[width=1.0\linewidth]{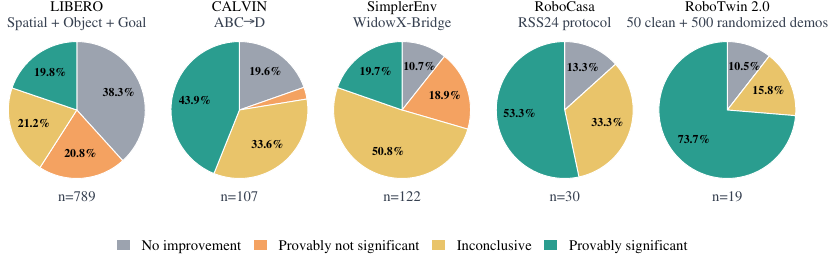}
    \caption{
    Reported gains from public scores alone. 
    Each pie sorts one benchmark's previous-best-to-new comparisons into the legend's four groups (bounds derived in Appendix~\ref{app:bounded-score-wald-test}). 
    Each comparison uses the previous best as reported in the new policy's own paper, rather than the highest earlier score we could find on arXiv.
    LIBERO pools three suites. 
    $n$ is the number of comparisons. 
    All bounds use a $5\%$ significance level.
    }
    \label{fig:statistical-significance-pies}
\end{figure}

\noindent\textbf{Method.}
We compare a new policy $B$ against a previously published policy $A$ with a lower aggregate score. Each benchmark consists of $T$ tasks, with $S$ paired samples per task. After evaluating both policies on the benchmark, a set of observed outcomes indexed by task and sample is collected, with $Y_{A,t,i},Y_{B,t,i}\in\{0,\dots,R\}$ for task $t\in{1,\dots,T}$ and sample $i\in{1,\dots,S}$. We assume that each policy is published with an average score \(\widehat\mu_A\), \(\widehat\mu_B\), where \(\widehat\mu_A < \widehat\mu_B \).

Define the empirical gap
$
\widehat\Delta:=\widehat\mu_B-\widehat\mu_A.
$
When both policies are evaluated on the same benchmark instances, we use a
one-sided paired Wald test with a task-stratified variance estimator. Let
\begin{equation*}
    \delta_{t,i}=Y_{B,t,i}-Y_{A,t,i},
    \qquad
    d_t=\sum_{i=1}^S \delta_{t,i},
    \qquad
    s_t=\sum_{i=1}^S \delta_{t,i}^2.
\end{equation*}
The paired stratified Wald statistic for testing
$
H_0:\Delta\leq 0
\text{ against }
H_1:\Delta>0,
$
where \(\Delta\) denotes the benchmark-average improvement in the top-line
metric, is
\begin{equation*}
    Z
    =
    \frac{TS(\widehat\mu_B - \widehat\mu_A)}
    {\sqrt{\frac{S}{S-1}\sum_{t=1}^T\left(s_t-d_t^2/S\right)}}.
\end{equation*}
%
%

The paired Wald statistic cannot be computed from top-line metrics alone.
Lacking access to the full paired outcome data, we derive two cutoffs on
the reported empirical gap. The rejection-feasibility threshold
$
\delta_{\exists}(\widehat\mu_A;T,S,R,\alpha)
$
is chosen so that if
$
\widehat\Delta
<
\delta_{\exists}(\widehat\mu_A;T,S,R,\alpha),
$
then no paired outcome table compatible with the reported top-line metrics
can reject the null hypothesis. The rejection-guarantee threshold
$
\delta_{\forall}(\widehat\mu_A;T,S,R,\alpha)
$
is chosen such that
$
\widehat\Delta
\geq
\delta_{\forall}(\widehat\mu_A;T,S,R,\alpha),
$
implies that every paired outcome table compatible with the reported top-line metrics
rejects the null hypothesis. The derivation is given in
Appendix~\ref{app:bounded-score-wald-test}.

For paired evaluations, these top-line bounds are sharp. Below the
feasibility threshold, rejection is impossible from the reported aggregate
scores. Above the guarantee threshold, rejection is forced by the reported
aggregate scores. Between the corresponding pointwise feasibility and
guarantee conditions, the aggregate scores alone do not determine
significance: the same reported counts are compatible with paired outcome
tables that reject and with paired outcome tables that do not reject.

Some audited papers do not evaluate both policies on the same benchmark
instances, choosing instead to evaluate on iid samples drawn from the benchmark distribution.
For these comparisons, the natural reference procedure is a stratified
two-sample Wald test rather than the paired Wald test above. We still use the
paired envelopes as conservative top-line cutoffs, but in such cases the thresholds will no longer be sharp. 

Figure~\ref{fig:statistical-significance-pies} shows the provably significant share is small on LIBERO and SimplerEnv and much larger on RoboCasa and RoboTwin~2.0. This is partly because RoboCasa and RoboTwin~2.0 have fewer reported results, so the scores sit further apart and larger gaps are easier to prove significant.


\vspace{-0.5em}
\section{Creeping Overfitting}
\label{sec:creeping-overfitting}
\vspace{-0.5em}

A benchmark has a fixed and usually narrow test distribution (for example, over object types, object poses, backgrounds, and camera poses). Its official test set is a set of finite samples drawn from that distribution.
Over time, models can become too tuned to the benchmark rather than acquiring broader manipulation skill, which prior work calls \emph{creeping overfitting}~\cite{Ponce2006DatasetIssues}.  Fitting to the narrow test distribution is \emph{distribution overfitting}, and fitting to the fixed samples is \emph{sample overfitting}.

\noindent\textbf{Distribution overfitting.}
CALVIN's most commonly used protocol, ABC$\to$D, trains on three table scenes (A, B, C) and tests on a fourth held-out scene D, where each evaluation episode is a chain of five language-conditioned tasks scored as the average number completed (ATC).  During training the table blocks can start at any pose in a range, but the released scene-D evaluation fixes them at the same poses. 
We resampled the block poses from that same training range and held everything else fixed. ATC fell by $1.03$ for X-VLA~\cite{Zheng2025XVLA}, $0.75$ for GR-1~\cite{Wu2023GR1}, and $0.50$ for RoboFlamingo~\cite{Li2024RoboFlamingo} with 5/5-chain drops of $25.0\%$, $13.5\%$, and $6.4\%$ (Figure~\ref{fig:diagnostic-teaser}c). 
Later policies show greater drops.

\begin{figure}[!t]
    \centering
    \includegraphics[width=0.95\linewidth]{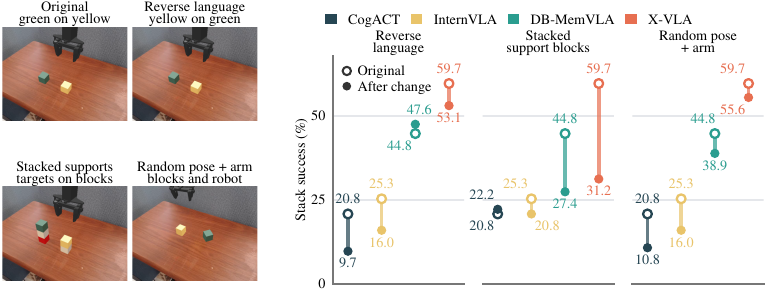}
    \caption{
    SimplerEnv stack task with three changes that stay inside the training distribution.  Left, rendered initial states for the original protocol and the three changes.  Right, original fixed-grid stack success and success after each change.
    }
    \label{fig:creeping-overfitting-simplerenv}
\end{figure}

On SimplerEnv, we changed the WidowX stack task in three ways (Figure~\ref{fig:creeping-overfitting-simplerenv}).  We measured CogACT-Base~\cite{Li2024CogACT}, InternVLA-M1~\cite{Chen2025InternVLAM1}, X-VLA-WidowX~\cite{Zheng2025XVLA}, and Dexbotic / DB-MemVLA~\cite{Xie2025Dexbotic} against a matched local calibration, our rerun on the official WidowX stack protocol (Appendix~\ref{app:experiment-details-creeping}).  We reversed the instruction (yellow on green instead of green on yellow), stacked the target block on a support block instead of the table, and resampled block positions and the initial arm pose.  All three changes already appear in the training set about as often as the official conditions do (Appendix~\ref{app:experiment-details-creeping}).  So they do not make the test more out-of-distribution than those conditions, and any drop shows overfitting rather than weak generalization.

The lower-scoring policies (CogACT-Base, InternVLA-M1) drop most on the reversed instruction, while the higher-scoring policies (X-VLA-WidowX, Dexbotic) hold there but drop sharply under the stacked-support change.  The benchmark uses a single fixed camera, which makes a block's depth and height hard to recover and may force policies to assume every block sits on the table. The stacking drop is consistent with that assumption.

\noindent\textbf{Sample overfitting.}
We redrew the fixed sample on LIBERO and CALVIN.  On LIBERO, redrawing the init-state file from the same generator with $10{,}000$ rollouts per policy moved success by less than a percentage point against a matched local calibration (Spatial Forcing~\cite{Li2025SpatialForcing} $-0.62$, SimVLA~\cite{Luo2026SimVLA} $+0.30$, \(\pi_{0.5}\)   LeRobot~\cite{PhysicalIntelligence2025Pi05,Cadene2026LeRobot} $-0.14$).  On CALVIN, drawing two fresh sequence manifests from the official generator moved ATC by at most $0.11$ for the CALVIN policies above (X-VLA, GR-1, and RoboFlamingo), an order of magnitude below the drops we saw when we resampled the poses.  Closed-loop rollouts are not bitwise deterministic (Appendix~\ref{app:bitwise-nondeterminism}), so each calibration carries noise that leaves the sign of any one policy's move uncertain. Either way the moves are small, no larger than that noise and far below the pose-resampling drops on the same CALVIN policies.


\section{Data-Source Dependence}
\label{sec:data-source-dependence}

A rising success rate is read across the field as evidence of a more capable policy.
This section asks when that reading can be inflated by the choice of training data rather than earned by capability.
The risk exists only when the training data sits far from the test, because then a high score looks like generalization across the gap.
For LIBERO, CALVIN, RoboCasa, and RoboTwin~2.0 the training data covers the test or the gap is small, so this diagnostic does not apply.

SimplerEnv is the exception.
Its test runs in a simulated WidowX setup, but the training data is real-world BridgeData V2 demonstrations~\cite{Walke2023BridgeDataV2} far from that test.
The gap is real, so a high score reads as generalization across it.


However, SimplerEnv does not restrict where training data comes from.
Without that restriction, the gap can be removed instead of crossed, by collecting training data right next to the test.
We show this with an extreme case.
X-VLA, a $0.9$B model with broad real and simulated pretraining, reports $95.8\%$ ($92/96$)~\cite{Zheng2025XVLA}, the field's reference for a near-solved score.
We train a separate $22$M policy per task, with no robotics pretraining, on $120$ scripted expert demonstrations recorded in simulation right beside the official test.
Together they reach $94.8\%$ ($91/96$).

We do not claim the small policy is better, only that on a benchmark with unrestricted training data a high score alone is not evidence of capability.
A genuinely strong method could reach the same score through real generalization, and the score alone cannot tell the two apart.
The fix needs to come from the benchmark itself, either by restricting the training set or by asking each submission to report how close its training data sits to the test.


\section{Discussion}
\label{sec:discussion}

All four failure modes share one root cause: a benchmark score is an empirical average of success over a fixed test distribution, treated as evidence of capability over the much broader distribution of interest. 
Each diagnostic catches one way the score can detach from the capability it stands in for: the test instances admit a shortcut, the empirical average is too noisy to resolve the claimed gap, the test distribution is a narrow slice of the variation we care about, or the gap between training and test data drives the score more than capability does.

\noindent\textbf{What a future manipulation benchmark should look like.}
Our diagnostics point to two design goals for future manipulation benchmarks.

First, the test distribution should cover broad variation in objects and scenes.
Our creeping-overfitting results on CALVIN and SimplerEnv, together with prior LIBERO-Pro and LIBERO-Plus findings on LIBERO~\cite{Zhou2025LIBEROPRO,Fei2025LIBEROPlus}, show that a high score on a narrow test distribution is not evidence that a policy performs at the same level under broad variation in objects and scenes. A wider distribution leaves less room for a brittle policy to score well without generalizing.
The shortcut diagnostic results align with this. The 0.09B DINO+MLP baseline matches SOTA on LIBERO and CALVIN but trails by more than 30\% on RoboTwin~2.0, whose object and scene variation is much greater. Difficulty should come from this variation, not from how far the training data is from test.  Section~\ref{sec:data-source-dependence} shows that a 22M policy matches a 0.9B foundation VLA simply by training on demonstrations collected next to the test, so a benchmark whose difficulty rests on that distance can be gamed.

Second, the official protocol should state what score gap counts as a statistically significant improvement over the previous SOTA, so that gains smaller than evaluation noise are not reported as progress.
The protocol should also use a test set large enough to detect the gains authors want to claim.
When the test set is fixed, the protocol should require each policy's per-instance outcomes, not only the aggregate score.
With those outcomes anyone can run the paired test, which can prove significance at far smaller gaps than the aggregate score allows.

\noindent\textbf{Until such benchmarks exist.}
The benchmarks we audit remain perfectly useful for everyday development work, such as fast iteration, debugging, and comparison against one's own earlier runs.
For shortcut solvability and data-source dependence, a high score is not by itself evidence of the capability the benchmark is taken to measure.
Creeping overfitting and weak significance weaken a reported result rather than invalidate it.
Either way, the result alone does not establish progress for the field.
Authors who think a commonly-reported benchmark fails one of our diagnostics can use this paper to support the argument that they should not be required to report on that benchmark, and that a benchmark they consider stronger is an acceptable substitute.
Reviewers can use it in the opposite direction, to question a small gain claimed as progress on such a benchmark.

\noindent\textbf{Beyond fixed-population averages.}
The diagnostics in this paper audit benchmarks that score a policy by its empirical average over a fixed test distribution.
A line of work asks whether that scoring rule can be replaced. Approaches based on item response theory model success as a function of a policy's effectiveness relative to the difficulty of a given instance~\cite{MartinezPlumed2019IRTAI,Rodriguez2021IRTLeaderboards}.
Combined with adaptive selection of test instances by this difficulty~\cite{WeissKingsbury1984CAT,SongFlach2021AdaptiveBenchmarks}, this direction could address creeping overfitting, data-source dependence, and statistical-power constraints jointly rather than one at a time.


\section{Limitations}
\label{sec:limitations}

The ideal test would train many policies on one shared set of demonstrations, evaluate them on a large test set that samples the capabilities we care about, and check whether the benchmark's policy ranking matches the ranking on that sample. The Introduction explains why that test is impractical. Our four diagnostics are a practical substitute.

For some diagnostics, like creeping overfitting, we could run only a few policies. Reproducing a policy needs both its open-source weights and the evaluation script that produced its reported numbers. Most policies do not release both, so we cannot reproduce them.


\section{Conclusion}
\label{sec:conclusion}

A benchmark score is useful only if it provides evidence of the capability we care about.
We identify four ways in which a high score can fail to be that evidence, and provide a diagnostic for each.
We ran the diagnostics on five widely used manipulation benchmarks.
The most-reported ones failed the most diagnostics.
We release the diagnostics so authors and reviewers can run them before reporting a score.



\bibliography{references}  


\clearpage
\appendix
\section{Experiment Details}
\label{app:experiment-details}

\begin{figure}[t]
    \centering
    \includegraphics[width=\linewidth]{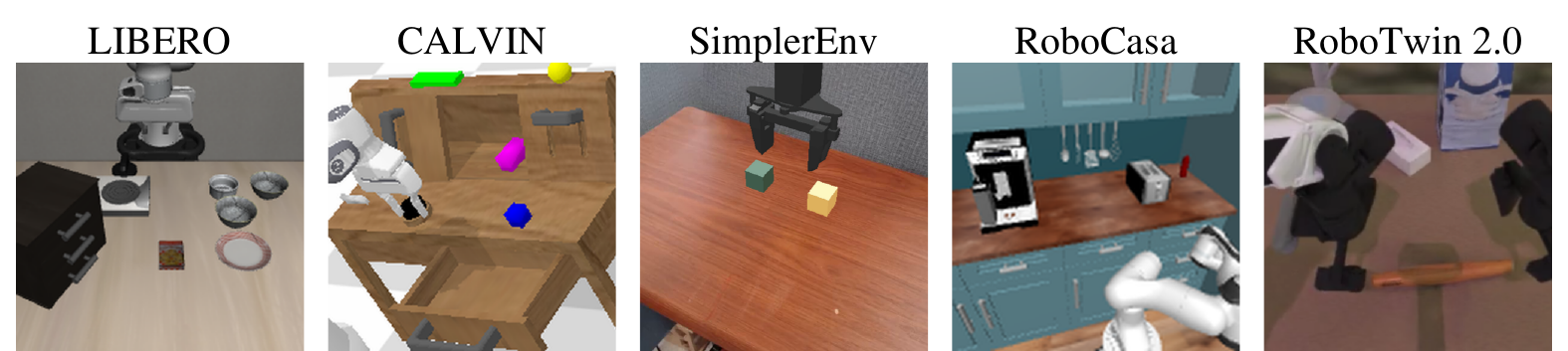}
    \caption{
    Example rendered observations from the five audited benchmarks, shown only to orient readers to the visual setup. From left to right, LIBERO~\cite{Liu2023LIBERO}, CALVIN~\cite{Mees2022CALVIN}, SimplerEnv~\cite{Li2025SimplerEnv}, RoboCasa~\cite{Nasiriany2024RoboCasa}, and RoboTwin~2.0~\cite{Chen2025RoboTwin2}.
    }
    \label{fig:benchmark-environment-examples}
\end{figure}

Figure~\ref{fig:benchmark-environment-examples} shows example observations from the five benchmarks we audit.

\subsection{Shortcut Solvability}
\label{app:experiment-details-shortcut}

\noindent\textbf{Probe.}
The probe is a DINOv2 image encoder~\cite{Oquab2023DINOv2} with a small MLP head, fine-tuned end to end, and it has no diffusion head and no action tokenizer. LIBERO, CALVIN, SimplerEnv, and RoboTwin~2.0 use a ViT-B/14 backbone, while RoboCasa uses the smaller ViT-S/14. Except on SimplerEnv (described below), the probe is never conditioned on language. It is told which task it faces only through a learned per-task embedding. Because the embedding is initialized at random and not taken from any language model, it carries none of the semantic content of a pretrained language embedding. LIBERO, CALVIN, and RoboTwin~2.0 use a multi-task model that takes this embedding, while RoboCasa trains a separate model per task and so needs none. The image inputs follow each benchmark's standard policy interface, the agentview and wrist cameras with proprioception on LIBERO, the static and gripper cameras on CALVIN, and three camera views at $128\times128$ on RoboCasa.

\noindent\textbf{Probe tuning and checkpoint selection.}
We train the probe separately for each LIBERO suite and each CALVIN protocol, sweeping learning rate, augmentation strength, and the training-step budget over a small grid, and we report the best of several checkpoints scored on the suite. Training one model per suite is standard practice, since OpenVLA fine-tunes separately on each task suite~\cite{Kim2025OpenVLA}, MoLA fine-tunes and evaluates per suite~\cite{Li2026MoLA}, and CORAL trains one expert per task~\cite{Luo2026CORAL}. The claim of Section~\ref{sec:shortcut-solvability} is that a shortcut exists, not that one method beats another, so we tune the probe deliberately to test whether any small probe can reach the score from the same inputs.

\noindent\textbf{Best checkpoint versus a fixed checkpoint.}
The score we report is the best of several checkpoints, each scored on the reported suite. The benchmark gives no separate validation set, so that suite is also the test set. Keeping the checkpoint that scores best on it is therefore selection on the test set. This is optimistic, but the benchmark allows it. This freedom is a weakness of the benchmark, not of our probe, and any submission could exploit it. We do so here and make it visible. Table~\ref{tab:checkpoint-optimism} puts each best-checkpoint score beside the final-checkpoint score, the score at the last training step with no selection. On LIBERO the two differ by $0.6$ to $4.4$ percentage points, with a final-checkpoint mean of $95.1\%$ against a best-checkpoint mean of $97.6\%$. On CALVIN they differ by at most $0.09$ average tasks completed, and on D$\to$D the reported score is already the final checkpoint.

\begin{table}[h]
\centering
\caption{Best-checkpoint and final-checkpoint probe scores. The best-checkpoint column is the best of all checkpoints scored on the reported suite and is what the main table reports. The final-checkpoint column is the score at the last training step, with no checkpoint selection. LIBERO values are success rate over $500$ trials per suite. CALVIN values are average tasks completed out of $5$ over $1000$ sequences.}
\label{tab:checkpoint-optimism}
\small
\begin{tabular}{@{}llrr@{}}
\toprule
Benchmark & Suite or protocol & Best checkpoint & Final checkpoint \\
\midrule
LIBERO & Spatial   & $99.0\%$  & $98.2\%$ \\
       & Object    & $100.0\%$ & $99.4\%$ \\
       & Goal      & $98.8\%$  & $94.8\%$ \\
       & Long      & $92.4\%$  & $88.0\%$ \\
\midrule
CALVIN & D$\to$D   & $3.123$   & $3.123$ \\
       & ABC$\to$D & $3.242$   & $3.149$ \\
\bottomrule
\end{tabular}
\end{table}

\noindent\textbf{LIBERO and CALVIN.}
On LIBERO we train one probe per suite on the official raw HDF5 demonstrations, evaluate under the official protocol with $50$ trials per task and $500$ per suite, and report success rate. On CALVIN we train one probe per protocol, condition on a $34$-way task id, and evaluate on the official $1000$-sequence long-horizon benchmark with episode length $360$, reporting average tasks completed. We report full evaluations for all three protocols, D$\to$D, ABC$\to$D, and ABCD$\to$D. The probe scores slightly higher on ABC$\to$D than on D$\to$D, which is surprising because D$\to$D trains and tests on the same scene. The cause is how CALVIN labels its training data. CALVIN cuts each demonstration into short language-labeled segments, and a segment often begins partway through the motion. For lifting a block, a segment can start with the gripper already above the block and contain only the final grasp. A probe conditioned on the task id alone then never sees the approach to the block during training and cannot perform it at test time. This artifact lowers the probe's D$\to$D score in particular, so D$\to$D falls below ABC$\to$D.

\noindent\textbf{RoboCasa.}
On RoboCasa~\cite{Nasiriany2024RoboCasa} we follow the RSS24 protocol. We train one probe per task on $50$ human demonstrations with AdamW for $1500$ epochs. Evaluation uses the RSS24 object-instance split B and the five fixed layout and style pairs $(1,1)$, $(2,2)$, $(4,4)$, $(6,9)$, and $(7,10)$, with no camera or texture randomization and $50$ rollouts per task. We report the strongest variant, a per-task model with an MLP head, pooled over the $24$ RSS24 manipulation tasks, $226/1200 = 18.8\%$. Multi-task and transformer-head variants scored lower.

\noindent\textbf{RoboTwin~2.0.}
For RoboTwin~2.0, we train the probe DINO+MLP policy following the evaluation protocol adopted by recent methods\cite{Yang2026ABotM0,MotuBrainTeam2026MotuBrain}. Specifically, we train a single multi-task policy using demonstrations from all 50 bimanual tasks. For each task, we use 50 demonstrations collected in the clean environment and 500 demonstrations collected in the randomized environment, resulting in 27,500 demonstrations in total. The clean environment uses a fixed clean background, a fixed table height, and no distractor objects on the table. The randomized environment introduces variations in the background, table height, and lighting, and includes multiple distractor objects on the table. To condition the policy on the target task, we assign each task a learned task embedding and concatenate it with the visual feature vector before feeding them into the MLP action head. For evaluation, we test the probe policy on all tasks under both clean and randomized settings, with 100 rollouts per task and setting, following the same evaluation protocol as prior methods. Table~\ref{tab:robotwin2_per_task_success_rates} reports the per-task success rates.

\begin{table}[h]
\centering
\caption{DINO+MLP probe policy per-task success rates on the RoboTwin~2.0 benchmark.}
\label{tab:robotwin2_per_task_success_rates}
\footnotesize
\setlength{\tabcolsep}{3.2pt}
\renewcommand{\arraystretch}{1.05}
\begin{tabularx}{\linewidth}{
>{\ttfamily\raggedright\arraybackslash}X
rr
@{\hspace{1.35em}}
>{\ttfamily\raggedright\arraybackslash}X
rr
}
\toprule
\multicolumn{1}{c}{\normalfont Task} & \multicolumn{1}{c}{\normalfont Clean} & \multicolumn{1}{c}{\normalfont Rand.} &
\multicolumn{1}{c}{\normalfont Task} & \multicolumn{1}{c}{\normalfont Clean} & \multicolumn{1}{c}{\normalfont Rand.} \\
\midrule
adjust\_bottle & 95.0\% & 90.0\% & place\_can\_basket & 50.0\% & 57.0\% \\
beat\_block\_hammer & 47.0\% & 42.0\% & place\_cans\_plasticbox & 97.0\% & 97.0\% \\
blocks\_ranking\_rgb & 44.0\% & 55.0\% & place\_container\_plate & 62.0\% & 51.0\% \\
blocks\_ranking\_size & 36.0\% & 40.0\% & place\_dual\_shoes & 24.0\% & 27.0\% \\
click\_alarmclock & 64.0\% & 68.0\% & place\_empty\_cup & 90.0\% & 87.0\% \\
click\_bell & 83.0\% & 81.0\% & place\_fan & 42.0\% & 43.0\% \\
dump\_bin\_bigbin & 48.0\% & 50.0\% & place\_mouse\_pad & 15.0\% & 20.0\% \\
grab\_roller & 100.0\% & 100.0\% & place\_object\_basket & 56.0\% & 62.0\% \\
handover\_block & 67.0\% & 67.0\% & place\_object\_scale & 50.0\% & 47.0\% \\
handover\_mic & 86.0\% & 80.0\% & place\_object\_stand & 79.0\% & 76.0\% \\
hanging\_mug & 8.0\% & 16.0\% & place\_phone\_stand & 58.0\% & 58.0\% \\
lift\_pot & 99.0\% & 100.0\% & place\_shoe & 69.0\% & 67.0\% \\
move\_can\_pot & 69.0\% & 61.0\% & press\_stapler & 99.0\% & 92.0\% \\
move\_pillbottle\_pad & 63.0\% & 59.0\% & put\_bottles\_dustbin & 9.0\% & 7.0\% \\
move\_playingcard\_away & 83.0\% & 86.0\% & put\_object\_cabinet & 50.0\% & 36.0\% \\
move\_stapler\_pad & 6.0\% & 5.0\% & rotate\_qrcode & 42.0\% & 44.0\% \\
open\_laptop & 96.0\% & 95.0\% & scan\_object & 82.0\% & 77.0\% \\
open\_microwave & 14.0\% & 21.0\% & shake\_bottle & 100.0\% & 100.0\% \\
pick\_diverse\_bottles & 83.0\% & 68.0\% & shake\_bottle\_horizontally & 100.0\% & 100.0\% \\
pick\_dual\_bottles & 97.0\% & 93.0\% & stack\_blocks\_three & 28.0\% & 31.0\% \\
place\_a2b\_left & 30.0\% & 24.0\% & stack\_blocks\_two & 62.0\% & 58.0\% \\
place\_a2b\_right & 25.0\% & 32.0\% & stack\_bowls\_three & 20.0\% & 25.0\% \\
place\_bread\_basket & 83.0\% & 80.0\% & stack\_bowls\_two & 60.0\% & 58.0\% \\
place\_bread\_skillet & 80.0\% & 69.0\% & stamp\_seal & 33.0\% & 29.0\% \\
place\_burger\_fries & 91.0\% & 94.0\% & turn\_switch & 46.0\% & 45.0\% \\
\midrule
\multicolumn{1}{l}{\normalfont\textbf{Average}} & \textbf{60.4\%} & \textbf{59.4\%} & \multicolumn{3}{c}{} \\
\bottomrule
\end{tabularx}
\end{table}

\noindent\textbf{SimplerEnv.}
SimplerEnv evaluates WidowX/Bridge policies, and its standard training source is the real-world BridgeData V2 demonstrations~\cite{Walke2023BridgeDataV2}. Unlike LIBERO and CALVIN, Bridge does not draw its instructions from a small fixed set, so the task-id lookup we use on those benchmarks does not apply here. We keep the same DINOv2 ViT-B/14 encoder and MLP head, but the model reads the instruction through a pretrained SigLIP text embedding alongside the image and proprioception. We train it on BridgeData V2 and evaluate on the official WidowX tasks of stack, carrot, spoon, and eggplant. The probe scores $0.0\%$. A small model trained on the standard Bridge data does not reach the SimplerEnv score, so unlike LIBERO the benchmark shows no shortcut of this kind.

\noindent\textbf{Policy sources for Table~\ref{tab:benchmark-audit-matrix}}: OpenVLA~\cite{Kim2025OpenVLA}; X-VLA~\cite{Zheng2025XVLA}, with the RoboTwin~2.0 X-VLA cell taken from the protocol-matched comparison reported by ABot-M0~\cite{Yang2026ABotM0}, because X-VLA's own RoboTwin~2.0 result does not train with 50 clean plus 500 randomized demonstrations per task; DeeR-VLA~\cite{Yue2024DeeRVLA}; RoboFlamingo~\cite{Li2024RoboFlamingo}; GR00T-N1~\cite{NVIDIA2025GR00TN1}; T-MEE~\cite{Bai2026ActionErrorDistributions}; MoH~\cite{Jing2025MixtureHorizons}; MoLA~\cite{Li2026MoLA}; CORAL$_{\text{SimVLA}}$~\cite{Luo2026CORAL}; MDT-V~\cite{Reuss2024MDT}; MMaDA-VLA~\cite{Liu2026MMaDAVLA}; Xiaomi-0~\cite{Cai2026XiaomiRobotics0}; MotuBrain~\cite{MotuBrainTeam2026MotuBrain}; X-WAM~\cite{Guo2026XWAM}.

\subsection{Creeping Overfitting}
\label{app:experiment-details-creeping}

This appendix gives the SimplerEnv details for Section~\ref{sec:creeping-overfitting}.

\paragraph{Setup.}
We evaluate four WidowX/Bridge policies on the SimplerEnv stack task: CogACT-Base, InternVLA-M1, X-VLA-WidowX, and Dexbotic (DB-MemVLA). All runs use the standard 60-step limit for this task.

\paragraph{Matched local calibration.}
The official WidowX stack test is a fixed grid of $24$ start states. Our calibration reruns that grid on our hardware, repeating each start state $12$ times, for $288$ episodes per policy. We report each change as the shift in stack success against this calibration on the same hardware, so hardware and sampling differences drop out and only the change remains.

\paragraph{The three changes.}
Each change holds the task, scene, camera, robot, and success check fixed, alters one factor, and runs $288$ episodes per policy.
\begin{itemize}
\item \emph{Reverse language.} The instruction asks for yellow on green instead of the official green on yellow.
\item \emph{Stacked support.} The two cubes start on one or two support blocks instead of directly on the table.
\item \emph{Random pose and arm.} The cube positions and the initial arm pose are redrawn at random instead of taken from the fixed grid.
\end{itemize}
These three belong to a larger run of seven conditions ($2{,}016$ episodes per policy) released with the paper.

\paragraph{Why the changes stay in distribution.}
The official test uses one fixed grid, but the BridgeData V2 training demonstrations~\cite{Walke2023BridgeDataV2}, recorded on a real WidowX robot, are not tied to that grid, so redrawn cube positions and arm poses are ordinary training states. Stacked blocks also appear in training. Across $53{,}192$ demonstrations, ``block'' appears in $869$, ``cube'' in $751$, and ``tower'' in $282$. Both color orders appear, the official green on yellow in $41$ demonstrations and the reversed yellow on green in $30$. Each change therefore stays close to the training data and is not much less common than the official condition.

\paragraph{Results.}
Table~\ref{tab:creeping-simplerenv} reports stack success for each policy on the calibration and the three changes.

\begin{table}[h]
\centering
\caption{SimplerEnv stack success rate (\%) on the matched local calibration and the three in-distribution changes, $288$ episodes per policy per column. Some calibration values differ from the numbers these policies report. This is expected, not an error. Several of these policies report under longer, easier episode step limits, while we hold every policy to the same shorter standard step limit.}
\label{tab:creeping-simplerenv}
\small
\begin{tabular}{@{}lrrrr@{}}
\toprule
Policy & Calibration & Reverse language & Stacked support & Random pose + arm \\
\midrule
CogACT-Base          & 20.8 & 9.7  & 22.2 & 10.8 \\
InternVLA-M1         & 25.3 & 16.0 & 20.8 & 16.0 \\
X-VLA-WidowX         & 59.7 & 53.1 & 31.3 & 55.6 \\
Dexbotic (DB-MemVLA) & 44.8 & 47.6 & 27.4 & 38.9 \\
\bottomrule
\end{tabular}
\end{table}

\paragraph{Confidence intervals.}
We report two-sided $95\%$ confidence intervals for the distribution-overfitting
and sample-overfitting drops, recomputed from the first-hand rollout CSV and JSON outputs.
Every drop uses
\[
\text{drop} = (\text{matched local calibration}) - (\text{altered or fresh result}),
\]
so a positive value means the altered or fresh test scored lower than the
matched local calibration and a negative value means it scored higher.

\paragraph{Interval methods.}
\emph{Binary success rates.} For the SimplerEnv and LIBERO success-rate drops we
report a Newcombe--Wilson interval for the difference of two proportions. For a
calibration count $x_c/n_c$ and an altered or fresh count $x_a/n_a$, write
$p_c = x_c/n_c$, $p_a = x_a/n_a$, and $\text{drop} = p_c - p_a$. We first form
Wilson score intervals $[L_c, U_c]$ and $[L_a, U_a]$ for the two proportions with
$z = z_{0.975} \approx 1.95996$. For a proportion $p = x/n$, the Wilson interval
is
\[
\text{center} = \frac{p + z^2/(2n)}{1 + z^2/n},
\qquad
\text{half} = \frac{z\,\sqrt{p(1-p)/n + z^2/(4n^2)}}{1 + z^2/n},
\]
\[
[L, U] = [\,\text{center} - \text{half},\ \text{center} + \text{half}\,].
\]
The Newcombe interval for $p_c - p_a$ is
\[
\text{lower} = \text{drop} - \sqrt{(p_c - L_c)^2 + (U_a - p_a)^2},
\qquad
\text{upper} = \text{drop} + \sqrt{(U_c - p_c)^2 + (p_a - L_a)^2}.
\]
This treats rollout rows as independent Bernoulli outcomes. For SimplerEnv it is
a row-level uncertainty summary for the local rollout protocol, and it does not
model the fact that the calibration repeats the same $24$ official grid
instances $12$ times.

\emph{CALVIN distribution overfitting.} The calibration and resampled-pose results share
the same canonical $1000$-sequence manifest, so we use a paired row-level Wald
interval over global sequence indices. With $d_i$ the per-sequence difference in
tasks completed,
\[
d_i = (\text{calibration count})_i - (\text{resampled count})_i,
\qquad
\text{drop} = \operatorname{mean}_i d_i,
\]
\[
\text{CI} = \text{drop} \pm z\,\frac{\operatorname{sd}(d_i)}{\sqrt{1000}}.
\]
The same paired calculation gives the 5/5 chain-success drops after converting
each sequence to the binary indicator $\mathbf{1}[\text{count} = 5]$.

\emph{CALVIN sample overfitting.} The calibration uses the official $1000$-sequence
manifest and the fresh result pools two independent fresh $1000$-sequence
manifests, so we use an independent two-sample Wald interval over per-sequence
tasks-completed counts in $\{0,1,2,3,4,5\}$,
\[
\text{drop} = \operatorname{mean}(\text{calibration}) - \operatorname{mean}(\text{fresh}),
\]
\[
\text{SE} = \sqrt{\frac{\operatorname{var}(\text{calibration})}{1000}
+ \frac{\operatorname{var}(\text{fresh})}{2000}},
\qquad
\text{CI} = \text{drop} \pm z\,\text{SE}.
\]

\paragraph{Distribution-overfitting drops.}
Table~\ref{tab:ci-simplerenv-layer1} reports the four-policy, three-change
SimplerEnv stack-task drops in percentage points, and
Table~\ref{tab:ci-calvin-layer1} reports the CALVIN resampled-pose drops, with
average tasks completed (ATC) in tasks out of $5$ and the 5/5 chain-success
drops in percentage points.

\begin{table}[h]
\centering
\caption{SimplerEnv distribution-overfitting drops. Calibration and altered columns are
success counts out of $288$ episodes per policy. Drop and 95\% CI are in
percentage points, with a positive drop meaning the altered test scored lower
than the matched local calibration.}
\label{tab:ci-simplerenv-layer1}
\small
\begin{tabular}{@{}llrrrc@{}}
\toprule
Policy & Change & Calibration & Altered & Drop & 95\% CI \\
\midrule
CogACT-Base          & Reverse language    & $60/288$  & $28/288$  & $+11.11$ & $[+5.26, +16.95]$ \\
CogACT-Base          & Stacked support     & $60/288$  & $64/288$  & $-1.39$  & $[-8.09, +5.33]$  \\
CogACT-Base          & Random pose + arm   & $60/288$  & $31/288$  & $+10.07$ & $[+4.13, +15.99]$ \\
InternVLA-M1         & Reverse language    & $73/288$  & $46/288$  & $+9.38$  & $[+2.76, +15.91]$ \\
InternVLA-M1         & Stacked support     & $73/288$  & $60/288$  & $+4.51$  & $[-2.38, +11.35]$ \\
InternVLA-M1         & Random pose + arm   & $73/288$  & $46/288$  & $+9.38$  & $[+2.76, +15.91]$ \\
X-VLA-WidowX         & Reverse language    & $172/288$ & $153/288$ & $+6.60$  & $[-1.49, +14.57]$ \\
X-VLA-WidowX         & Stacked support     & $172/288$ & $90/288$  & $+28.47$ & $[+20.46, +35.96]$ \\
X-VLA-WidowX         & Random pose + arm   & $172/288$ & $160/288$ & $+4.17$  & $[-3.88, +12.14]$ \\
Dexbotic (DB-MemVLA) & Reverse language    & $129/288$ & $137/288$ & $-2.78$  & $[-10.84, +5.33]$ \\
Dexbotic (DB-MemVLA) & Stacked support     & $129/288$ & $79/288$  & $+17.36$ & $[+9.54, +24.89]$ \\
Dexbotic (DB-MemVLA) & Random pose + arm   & $129/288$ & $112/288$ & $+5.90$  & $[-2.14, +13.84]$ \\
\bottomrule
\end{tabular}
\end{table}

\begin{table}[h]
\centering
\caption{CALVIN distribution-overfitting drops on the shared $1000$-sequence manifest.
ATC drops are in tasks out of $5$; 5/5 chain-success drops are in percentage
points. A positive drop means the resampled-pose test scored lower than the calibration.}
\label{tab:ci-calvin-layer1}
\small
\begin{tabular}{@{}llrrrc@{}}
\toprule
Policy & Metric & Calibration & Resampled & Drop & 95\% CI \\
\midrule
X-VLA        & ATC               & $4.165$    & $3.138$    & $+1.027$ & $[+0.890, +1.164]$ \\
X-VLA        & 5/5 chain success & $709/1000$ & $459/1000$ & $+25.00$ & $[+21.29, +28.71]$ \\
GR-1         & ATC               & $3.244$    & $2.495$    & $+0.749$ & $[+0.624, +0.874]$ \\
GR-1         & 5/5 chain success & $431/1000$ & $296/1000$ & $+13.50$ & $[+10.23, +16.77]$ \\
RoboFlamingo & ATC               & $2.367$    & $1.869$    & $+0.498$ & $[+0.372, +0.624]$ \\
RoboFlamingo & 5/5 chain success & $209/1000$ & $145/1000$ & $+6.40$  & $[+3.57, +9.23]$   \\
\bottomrule
\end{tabular}
\end{table}

\paragraph{Sample-overfitting drops.}
Table~\ref{tab:ci-libero-layer2} reports the LIBERO fresh-init-state drops in
percentage points, and Table~\ref{tab:ci-calvin-layer2} reports the CALVIN
fresh-sequence drops, where the fresh column pools the two predeclared fresh
$1000$-sequence manifests and ATC is in tasks out of $5$.

\begin{table}[h]
\centering
\caption{LIBERO sample-overfitting drops. Calibration is success counts out of $2000$
episodes per policy; the fresh sample redraws the init-state file from the same
generator with $10{,}000$ rollouts per policy. Drop and 95\% CI are in
percentage points.}
\label{tab:ci-libero-layer2}
\small
\begin{tabular}{@{}llrrrc@{}}
\toprule
Policy & Metric & Calibration & Fresh sample & Drop & 95\% CI \\
\midrule
Spatial Forcing      & success rate & $1956/2000$ & $9718/10000$ & $+0.62$ & $[-0.18, +1.27]$ \\
SimVLA               & success rate & $1946/2000$ & $9760/10000$ & $-0.30$ & $[-1.15, +0.40]$ \\
$\pi_{0.5}$ / LeRobot & success rate & $1951/2000$ & $9741/10000$ & $+0.14$ & $[-0.69, +0.82]$ \\
\bottomrule
\end{tabular}
\end{table}

\begin{table}[h]
\centering
\caption{CALVIN sample-overfitting drops. The fresh sample pools two predeclared
independent fresh $1000$-sequence manifests. ATC is in tasks out of $5$, with a
positive drop meaning the fresh sample scored lower than the calibration.}
\label{tab:ci-calvin-layer2}
\small
\begin{tabular}{@{}llrrrc@{}}
\toprule
Policy & Metric & Calibration & Fresh sample & Drop & 95\% CI \\
\midrule
X-VLA        & ATC & $4.165$ & $4.1795$ & $-0.0145$ & $[-0.1287, +0.0997]$ \\
GR-1         & ATC & $3.244$ & $3.1370$ & $+0.1070$ & $[-0.0373, +0.2513]$ \\
RoboFlamingo & ATC & $2.367$ & $2.2980$ & $+0.0690$ & $[-0.0677, +0.2057]$ \\
\bottomrule
\end{tabular}
\end{table}

\paragraph{Confidence-interval source artifacts.}
The intervals were recomputed directly from the per-episode and per-sequence
rollout outputs rather than from summary notes. The SimplerEnv drops use the
$288$-episode fixed-grid calibration and the three altered-condition rollout outputs. The
CALVIN drops use the official $1000$-sequence aggregates for the calibration and
the resampled-pose test, and the two predeclared fresh $1000$-sequence aggregates for the
sample-overfitting experiment. The LIBERO drops use the official per-suite calibration and the
$10{,}000$-rollout fresh-init-state outputs, with a small number of cells taken
from documented retry reruns.

\subsection{Data-Source Dependence}
\label{app:experiment-details-data-source}

This appendix gives the SimplerEnv details for Section~\ref{sec:data-source-dependence}. The diagnostic is intentionally narrow: SimplerEnv evaluates policies in a simulated WidowX setup, while high benchmark scores are often read as evidence that a policy transfers from real BridgeData V2 demonstrations~\cite{Walke2023BridgeDataV2} to that simulator. Because SimplerEnv does not restrict training-data sources, we can instead train on scripted demonstrations generated in the same simulator around the official evaluation grid. This is not a claim that scripted demonstrations are comparable to real Bridge demonstrations; it tests whether the score itself identifies transfer across the data-source gap.

\paragraph{Protocol.}
We use the official SimplerEnv WidowX evaluation grid for four Bridge-style tasks: stack, carrot, spoon, and eggplant. Each task has $24$ fixed grid instances, for $96$ closed-loop evaluations in total. Training data are generated separately for each task, and evaluation runs on grid ids $0$--$23$ without the training jitter, using the benchmark task definitions, observations, episode step limits, and success checks. We report counts as well as percentages because each task has only $24$ trials.

\paragraph{Scripted demonstrations.}
For each task we collect $120$ successful scripted expert demonstrations in simulation, exactly $5$ per official grid id. The demonstrations use small object perturbations around the official grid, with $0.0025\,\mathrm{m}$ XY jitter and up to $15^\circ$ yaw jitter, and include terminal open/place padding so the final release behavior is represented in the behavior-cloning targets. The carrot dataset additionally canonicalizes gripper yaw modulo $\pi$. The generator uses the public official grid ids only as centers for the jittered demonstration distribution. This deliberately benchmark-adjacent setup tests whether unrestricted training data can avoid, rather than solve, the Bridge-to-SimplerEnv data-source gap.

\paragraph{Policy and training.}
We train one task-specific DINOv2-S + MLP behavior-cloning policy per task, initialized with a pretrained DINOv2-S visual encoder. Each policy is trained as a separate BC run on that task's scripted demonstrations. The visual encoder is fine-tuned, image augmentation is disabled, proprioception is included, actions are trained in the SimplerEnv environment-action convention with dataset-standard normalization, and deployment uses $5$-step action chunks with replanning every $5$ steps. Loader-side terminal-action padding is enabled for training. There is no checkpoint selection.
These runs are not an unbiased success-rate estimate: generator fixes were selected to demonstrate the existence of a high-scoring policy on the fixed grid.

\begin{table}[h]
\centering
\caption{SimplerEnv WidowX data-source-dependence results. Each task row uses a separate \mbox{DINOv2-S + MLP} policy trained only on $120$ scripted demonstrations generated in the same simulator around the official evaluation grid; the final row aggregates all four tasks. Evaluation uses the official $24$-instance grid for each task.}
\label{tab:data-source-simplerenv}
\small
\begin{tabular}{@{}lrrr@{}}
\toprule
Task & Scripted demos & Eval successes & Success rate \\
\midrule
Stack    & $120$ & $24/24$ & $100.0\%$ \\
Carrot   & $120$ & $23/24$ & $95.8\%$ \\
Spoon    & $120$ & $21/24$ & $87.5\%$ \\
Eggplant & $120$ & $23/24$ & $95.8\%$ \\
\midrule
All four & $480$ & $91/96$ & $94.8\%$ \\
\bottomrule
\end{tabular}
\end{table}

\paragraph{Interpretation.}
The result in Table~\ref{tab:data-source-simplerenv} is a fixed-grid existence result, not an unbiased success-rate estimate and not a controlled comparison to X-VLA. It shows that, when training data is unrestricted, a $91/96$ ($94.8\%$) score on the official fixed grid can be achieved by removing the data-source gap rather than crossing it. Therefore the score alone cannot distinguish the two explanations.

\section{Bitwise Nondeterminism in Policy Rollouts}
\label{app:bitwise-nondeterminism}

\subsection{Reproduction Targets}
\label{app:bitwise-targets}

The strictest reproduction target is bitwise determinism: the same checkpoint,
seed, initial state, wrapper, normalization, action postprocessing, simulator,
and evaluation code would produce bitwise-identical states, observations,
actions, success labels, and final score.  This appendix explains why bitwise
determinism should not be used as the sole cross-hardware criterion for
independent closed-loop manipulation evaluations.  A recent technical discussion
of LLM inference nondeterminism emphasizes that bitwise reproducibility depends
on the chosen system boundary and numerical implementation
details~\cite{He2025Nondeterminism}.  Here ``portable'' means applicable across
documented but not necessarily identical execution stacks, such as different CPU
or GPU hardware.

We separate four increasingly less exacting targets:
\begin{enumerate}
    \item \textbf{Bitwise determinism}: every simulator state, rendered
    observation, policy input, action, reward, done flag, and success label is
    bitwise identical at every step.
    \item \textbf{Per-episode outcome agreement}: per-episode success or failure
    labels match, even if hidden states or pixels differ.
    \item \textbf{Aggregate-score agreement}: the final success rate or CALVIN
    average-tasks-completed (ATC) score matches, possibly without matching the
    same episodes.
    \item \textbf{Protocol and statistical reproducibility}: the independent run
    follows the documented benchmark protocol, records per-episode outcomes when
    possible, and gives aggregate behavior compatible with finite-rollout
    uncertainty.
\end{enumerate}

Our paper claims use the last standard, while treating bitwise determinism as a
useful diagnostic.  This does not relax artifact standards: exact commands,
configs, checkpoints, seeds, initial states, wrappers, normalization,
postprocessing, success parsing, hardware, driver, CUDA, PyTorch, simulator, and
environment metadata are all part of the reproducible object.

\subsection{Diagnostic Evidence}
\label{app:bitwise-diagnostics-section}

Closed-loop evaluation makes bitwise determinism a fragile requirement.  A
manipulation benchmark does not merely apply a fixed action log to a simulator.
The simulator and policy form a discrete-time dynamical system: the simulator
state determines rendered images and proprioception; those observations
determine the next policy action; the next action changes contacts and future
states.  Tiny numerical differences from a simulator step or policy forward pass
can therefore remain invisible for many steps, or they can be amplified by
contact and feedback into different pixels, different actions, and different
success labels.  Executing a fixed action log open-loop is not sufficient
evidence for this question because it removes the policy-simulator feedback loop
that benchmark evaluation is meant to measure.

\begin{figure}[H]
    \centering
    \includegraphics[width=0.92\linewidth]{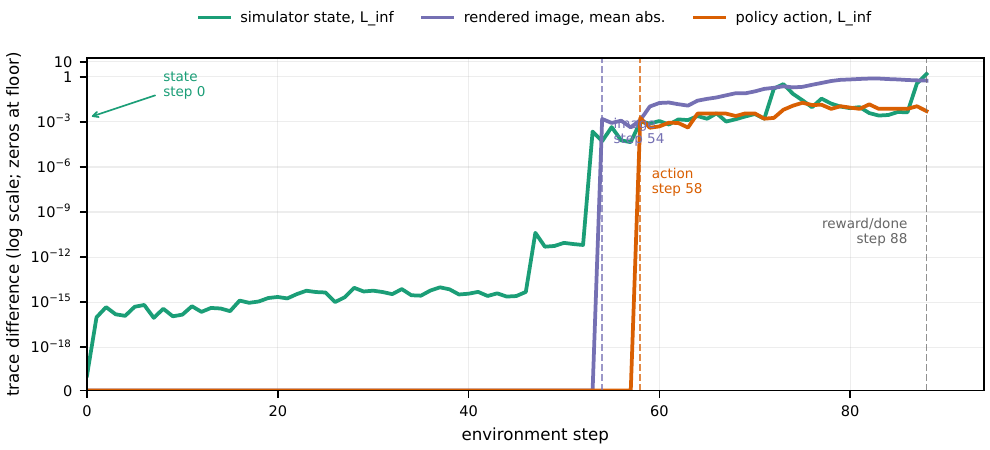}
    \caption{Closed-loop propagation in one LIBERO-Goal task 8 hardware
    comparison.  With policy, seed, initial state, wrapper, and rollout code
    fixed, changing CPU hardware while keeping the GPU class fixed first changes
    post-step simulator state at step 0; the difference reaches rendered images
    at step 54, actions at step 58, and reward/done at step 88.  State and
    action curves use \(L_\infty\) distance; the image curve uses mean absolute
    pixel difference.  Zeros are drawn at the log-scale floor.}
    \label{fig:bitwise-libero-task8-propagation}
\end{figure}

Figure~\ref{fig:bitwise-libero-task8-propagation} illustrates the mechanism:
the first bitwise difference is tiny and internal to the simulator, but later
enters pixels, actions, and the success parser through the feedback loop.  In
this task, contact instrumentation also localized the growth around
contact/constraint dynamics involving the bowl, including a bowl-table
contact-mode mismatch during the grasp/lift phase.

\begin{table}[H]
    \centering
    \caption{Diagnostic closed-loop trace-comparison tests.  Each row is one
    CPU or GPU hardware contrast for one benchmark stack.  Same-node controls
    were bitwise identical in the recorded arrays; the goal is not to survey
    all benchmark stacks, but to show that bitwise determinism can fail even
    when the benchmark protocol is fixed.}
    \label{tab:bitwise-diagnostics}
    \scriptsize
    \setlength{\tabcolsep}{4pt}
    \renewcommand{\arraystretch}{1.15}
    \begin{tabular}{@{}p{0.20\linewidth}p{0.25\linewidth}p{0.32\linewidth}p{0.14\linewidth}@{}}
        \toprule
        Stack & Hardware contrast & Observed propagation & Bitwise identical? \\
        \midrule
        LIBERO / OpenVLA-OFT &
        CPU: AMD EPYC 74F3 + RTX A6000 versus Intel Xeon Silver 4210R + RTX A6000 &
        \(10/10\) tasks diverged in simulator state; \(5/10\) reached pixels and
        actions. &
        No \\
        \addlinespace[0.25em]
        LIBERO / OpenVLA-OFT &
        GPU: same AMD EPYC 74F3 CPU model, RTX A6000 versus RTX 6000 Ada &
        Raw and processed actions diverged at step 10, followed by simulator
        state, image, and reward/done divergence. &
        No \\
        \addlinespace[0.25em]
        CALVIN / GR-1 &
        CPU: AMD-CPU node with RTX A6000 versus Intel-CPU node with RTX A6000 &
        \(0/50\) official sequence pairs diverged in this diagnostic sample. &
        Yes \\
        \addlinespace[0.25em]
        CALVIN / GR-1 &
        GPU: same AMD EPYC 74F3 CPU model, RTX A6000 versus RTX 6000 Ada &
        Raw and processed actions diverged at step 0, followed by simulator
        state and image-hash divergence. &
        No \\
        \addlinespace[0.25em]
        SimplerEnv / CogACT &
        CPU: AMD-CPU node with RTX A6000 versus Intel-CPU node with RTX A6000 &
        Actions and post-step simulator state diverged at step 0; images at
        step 3; reward/done/success at step 38. &
        No \\
        \addlinespace[0.25em]
        SimplerEnv / CogACT &
        GPU: same AMD EPYC 74F3 CPU model, RTX A6000 versus RTX 6000 Ada &
        Actions and simulator state diverged from step 0, followed by image
        divergence; both traces still ended unsuccessfully. &
        No \\
        \addlinespace[0.25em]
        RoboCasa / StarVLA &
        CPU: AMD-CPU node with RTX A6000 versus Intel-CPU node with RTX A6000 &
        Simulator and contact traces diverged immediately; image/action
        divergence followed at step 4 and reward/success at step 19. &
        No \\
        \addlinespace[0.25em]
        RoboCasa / StarVLA &
        GPU: same AMD EPYC 74F3 CPU model, RTX A6000 versus RTX 6000 Ada &
        Actions, contact traces, simulator state, and images diverged from the
        first steps; reward/success diverged at step 18. &
        No \\
        \bottomrule
    \end{tabular}
\end{table}

The LIBERO CPU contrast is the clearest positive case: with GPU class held
fixed as RTX A6000, changing CPU hardware broke bitwise determinism and
sometimes changed the final outcome.  The same-node controls and the CALVIN CPU
screen bound the claim in the other direction.  Bitwise determinism can hold; it
is just not a property that should be assumed across independent hardware
stacks.

\subsection{Implications for This Paper}
\label{app:bitwise-implications}

This distinction matters for paper sections that rely on rerunning documented
benchmark protocols.  For statistical significance, the relevant object is the
per-instance success/failure table under a documented protocol, because that is
what determines the paired uncertainty in
Section~\ref{sec:statistical-significance}.  Recovering the exact decimal in a
published aggregate table would not by itself establish a significant
improvement.  Conversely, a small aggregate mismatch does not by itself refute
reproduction when protocol fidelity is established and the discrepancy is
statistically compatible.

For shortcut solvability, the relevant claim is also protocol-level: under
official evaluation protocols, simple DINO+MLP policies with task-ID inputs
reach high benchmark scores on LIBERO and CALVIN.  The inference failure is that
those benchmark scores do not identify language understanding or large-scale VLA
pretraining.

The practical implication is to strengthen, not relax, the artifact standard.
Authors should release enough information for a reader to audit protocol
fidelity and to interpret statistically compatible discrepancies: checkpoint
references, code revisions, commands, configs, seeds, initial states, wrappers,
normalization, action postprocessing, success parsing, CPU and GPU metadata,
driver, CUDA, PyTorch, simulator versions, environment identifiers, and
per-episode outcome files.  Exact bitwise determinism is valuable when it is
available, but it should not be required as a cross-hardware pass/fail criterion
when protocol fidelity, per-episode outcomes, and statistical uncertainty are
reported.

\section{Top-Line Cutoffs for Bounded-Score Wald Tests}
\label{app:bounded-score-wald-test}

This appendix derives top-line-only cutoffs for bounded-score benchmark
comparisons. The primary derivation is for paired evaluations, where both
policies are evaluated on the same benchmark instances. We then show that the
paired cutoffs remain conservative for independently sampled evaluations with
the same task structure and the same number of samples per task.

Binary success/failure benchmarks correspond to \(R=1\). CALVIN corresponds
to \(R=5\), \(T=1\), and \(S=1000\).

\subsection{Paired Task-Stratified Wald Rule}

We compare two policies on a fixed benchmark with \(T\) tasks and \(S\)
paired samples per task. The tasks are treated as fixed strata, and the
paired samples within each task are assumed independent. Unstratified
benchmarks are represented by setting \(T=1\). Assume \(S>1\) and
\[
Y_{A,t,i},Y_{B,t,i}\in\{0,1,\ldots,R\}.
\]
For task \(t\) and sample \(i\), define the paired score difference
\[
\delta_{t,i}=Y_{B,t,i}-Y_{A,t,i}\in\{-R,\ldots,R\}.
\]
Let
\[
\Delta_t=\mathbb{E}[\delta_{t,i}]
=
\mu_{B,t}-\mu_{A,t},
\qquad
\sigma_t^2=\operatorname{Var}(\delta_{t,i}),
\]
and define the fixed-task estimand
\[
\Delta
=
\frac{1}{T}\sum_{t=1}^T \Delta_t.
\]
Because every task has the same number \(S\) of samples, the corresponding
empirical estimator is equal to the difference in top-line empirical metrics:
\[
\widehat\Delta_t=\frac{1}{S}\sum_{i=1}^S\delta_{t,i},
\qquad
\widehat\Delta=\frac{1}{T}\sum_{t=1}^T\widehat\Delta_t.
\]
For fixed \(T\) and \(S\to\infty\), the central limit theorem gives
\[
\frac{\widehat{\Delta}-\Delta}
{
\sqrt{
\sum_{t=1}^T \frac{\sigma_t^2}{T^2S}
}
}
\overset{d}{\longrightarrow}
\mathcal{N}(0,1),
\]
provided the limiting variance is nonzero. For finite \(S\), we use the
resulting Wald procedure as a nominal large-sample decision rule.

Define
\[
d_t=\sum_{i=1}^S \delta_{t,i},
\qquad
s_t=\sum_{i=1}^S \delta_{t,i}^2.
\]
The unbiased sample variance of the paired differences on task \(t\) is
\[
\widehat\sigma_t^2
=
\frac{1}{S-1}
\left(
s_t-\frac{d_t^2}{S}
\right).
\]
Therefore
\[
\widehat V
=
\sum_{t=1}^T\frac{\widehat\sigma_t^2}{T^2S}
=
\frac{1}{T^2S(S-1)}
\sum_{t=1}^T
\left(
s_t-\frac{d_t^2}{S}
\right),
\]
and, when \(\widehat V>0\),
\[
Z
=
\frac{\widehat\Delta}{\sqrt{\widehat V}}
=
\frac{
\sum_{t=1}^T d_t
}{
\sqrt{
\frac{S}{S-1}
\sum_{t=1}^T
\left(
s_t-\frac{d_t^2}{S}
\right)
}
}.
\]

For the finite-sample decision rule, we use the following convention when
\(\widehat V=0\):
\[
Z_*=
\begin{cases}
\widehat\Delta/\sqrt{\widehat V},
& \widehat V>0,\\
+\infty,
& \widehat V=0\text{ and }\widehat\Delta>0,\\
0,
& \widehat V=0\text{ and }\widehat\Delta=0,\\
-\infty,
& \widehat V=0\text{ and }\widehat\Delta<0.
\end{cases}
\]
For testing
\[
H_0:\Delta\leq 0
\qquad
\text{against}
\qquad
H_1:\Delta>0,
\]
the nominal one-sided Wald rule rejects when
\[
Z_*>z_{1-\alpha}.
\]
Throughout the top-line bounds below we assume \(0<\alpha<1/2\), so that
\(z_{1-\alpha}>0\).

For binary success/failure outcomes, \(R=1\) and
\(\delta_{t,i}^2=1\) exactly when policies \(A\) and \(B\) recorded different
outcomes. Thus \(s_t\) is the task-level discordance count, and the statistic
reduces to the usual paired stratified Wald statistic for binary outcomes.

\subsection{Top-Line Count Notation}

Suppose only the top-line empirical metrics are reported. Let
\[
N=TS,
\qquad
A=N\widehat\mu_A,
\qquad
B=N\widehat\mu_B,
\qquad
L=B-A,
\]
and assume in this subsection that
\[
A,B\in\{0,\ldots,RN\},
\qquad
B>A.
\]
The observed empirical gap is
\[
g=\widehat\mu_B-\widehat\mu_A=\frac{L}{N}.
\]
For fixed \(A\), the feasible positive count gaps are
\[
\mathcal L_A=\{1,\ldots,RN-A\}.
\]
For each \(L\in\mathcal L_A\), the corresponding top-line count for policy
\(B\) is \(B=A+L\).

The reported top-line counts imply
\[
\sum_{t=1}^T d_t=L.
\]
The paired Wald denominator depends on the paired outcome table only through
\[
Q
=
\sum_{t=1}^T
\left(
s_t-\frac{d_t^2}{S}
\right).
\]
Define
\[
c_\alpha
=
z_{1-\alpha}
\sqrt{\frac{S}{S-1}}.
\]
For fixed reported counts \(A\) and \(B=A+L\), the numerator of the Wald
statistic is fixed at \(L\). Since \(L>0\), the zero-variance convention above
implies that a compatible paired outcome table rejects if and only if
\[
L>c_\alpha\sqrt{Q}.
\]
This equivalence includes the case \(Q=0\), in which the displayed inequality
reduces to \(L>0\).

\subsection{Exact Lower Envelope for Paired Evaluations}

We first minimize \(Q\) over all paired outcome tables compatible with the
top-line counts. For any integer \(x\), write
\[
x=qS+\rho_S(x),
\qquad
q\in\mathbb{Z},
\qquad
\rho_S(x)\in\{0,\ldots,S-1\}.
\]
For a fixed task total \(d_t\), the integer differences
\(\delta_{t,1},\ldots,\delta_{t,S}\) minimize
\(\sum_i\delta_{t,i}^2\), subject to \(\sum_i\delta_{t,i}=d_t\), by being as
equal as possible. Hence every compatible paired outcome table satisfies
\[
s_t-\frac{d_t^2}{S}
\ge
\rho_S(d_t)-\frac{\rho_S(d_t)^2}{S}.
\]
Let
\[
\psi(r)=r-\frac{r^2}{S},
\qquad
0\le r<S.
\]
For \(x,y\in\{0,\ldots,S-1\}\),
\[
\psi(x)+\psi(y)\ge \psi((x+y)\bmod S).
\]
Indeed, if \(x+y<S\), the difference is \(2xy/S\); if \(x+y\ge S\), the
difference is \(2(S-x)(S-y)/S\). Repeatedly combining residuals across tasks
therefore gives a lower bound depending only on the total count gap \(L\).
Write
\[
L=qS+r,
\qquad
q=\left\lfloor \frac{L}{S}\right\rfloor,
\qquad
0\le r<S.
\]
Then every compatible paired outcome table satisfies
\[
Q
\ge
Q_{\mathrm{lo}}(L;T,S),
\]
where
\[
Q_{\mathrm{lo}}(L;T,S)
=
r-\frac{r^2}{S}.
\]

This lower bound is sharp. Choose integers
\(c_1,\ldots,c_T\in\{0,\ldots,R\}\) such that
\[
\sum_{t=1}^T c_t=q.
\]
If \(r>0\), then \(q<RT\), so choose a task \(t_0\) with \(c_{t_0}<R\). Define
paired differences by
\[
\delta_{t,i}=c_t
\quad
\text{for all }t,i,
\]
and, when \(r>0\), add one additional unit to \(r\) samples in task \(t_0\):
\[
\delta_{t_0,i}=c_{t_0}+1
\quad
\text{for }i=1,\ldots,r.
\]
The total difference is \(L\), and all task-level contributions to \(Q\) are
zero except the residual task, whose contribution is
\[
r-\frac{r^2}{S}.
\]
Thus the constructed difference table attains \(Q_{\mathrm{lo}}(L;T,S)\).

It remains to realize the specified top-line count \(A\). The total remaining
capacity under the constructed nonnegative differences is
\[
\sum_{t=1}^T\sum_{i=1}^S (R-\delta_{t,i})
=
RN-L
\ge A,
\]
because \(B=A+L\le RN\). Choose integer baseline scores
\(X_{t,i}\in\{0,\ldots,R-\delta_{t,i}\}\) summing to \(A\), and set
\[
Y_{A,t,i}=X_{t,i},
\qquad
Y_{B,t,i}=X_{t,i}+\delta_{t,i}.
\]
This gives a compatible paired outcome table with
\[
Q=Q_{\mathrm{lo}}(L;T,S).
\]
Therefore \(Q_{\mathrm{lo}}\) is the exact minimum of \(Q\) over compatible
paired outcome tables.

Consequently, rejection is impossible whenever
\[
L
\le
c_\alpha
\sqrt{
Q_{\mathrm{lo}}(L;T,S)
}.
\]
Conversely, if the inequality is reversed, then the sharpness construction
above gives a compatible paired outcome table that rejects.

The count-scale rejection-feasibility cutoff is
\[
L_{\exists}(A;T,S,R,\alpha)
=
\min
\left\{
L\in\mathcal L_A:
L
>
c_\alpha
\sqrt{
Q_{\mathrm{lo}}(L;T,S)
}
\right\},
\]
with the convention that the minimum is \(+\infty\) if the set is empty. The
normalized cutoff is
\[
\delta_{\exists}(\widehat\mu_A;T,S,R,\alpha)
=
\frac{L_{\exists}(A;T,S,R,\alpha)}{N}.
\]
Thus
\[
g
<
\delta_{\exists}(\widehat\mu_A;T,S,R,\alpha)
\]
implies that rejection is impossible from the reported top-line metrics.

Since
\[
Q_{\mathrm{lo}}(L;T,S)
=
L-\frac{L^2}{S}
\qquad
\text{for }1\le L\le S,
\]
the first count gap satisfying the feasibility inequality is
\[
\ell_{\exists}^{\star}(S,\alpha)
=
1+
\left\lfloor
\frac{z_{1-\alpha}^2 S}
{S-1+z_{1-\alpha}^2}
\right\rfloor.
\]
Therefore
\[
L_{\exists}(A;T,S,R,\alpha)
=
\begin{cases}
\ell_{\exists}^{\star}(S,\alpha),
&
\text{if }\ell_{\exists}^{\star}(S,\alpha)\le RN-A,\\
+\infty,
&
\text{otherwise.}
\end{cases}
\]

\subsection{Exact Upper Envelope for Paired Evaluations}

We next maximize \(Q\) over all paired outcome tables compatible with the
top-line counts. For one task, suppose the task has total score \(a\) for
policy \(A\) and total score \(b\) for policy \(B\). Define
\[
M_{\max}^{(R)}(a,b;S)
=
\max_{\{n_{uv}\}}
\sum_{u=0}^R\sum_{v=0}^R
n_{uv}(v-u)^2
\]
subject to
\[
n_{uv}\in\mathbb{Z}_{\ge 0},
\]
\[
\sum_{u=0}^R\sum_{v=0}^R n_{uv}=S,
\]
\[
\sum_{u=0}^R\sum_{v=0}^R u\,n_{uv}=a,
\qquad
\sum_{u=0}^R\sum_{v=0}^R v\,n_{uv}=b.
\]
Here \(n_{uv}\) is the number of samples in the task on which policy \(A\)
receives score \(u\) and policy \(B\) receives score \(v\). The corresponding
maximum task-level contribution to \(Q\), conditional on \((a,b)\), is
\[
q_{\max}^{(R)}(a,b;S)
=
M_{\max}^{(R)}(a,b;S)
-
\frac{(b-a)^2}{S}.
\]

The integer program above is the cleanest definition. The implementation uses
the following equivalent closed form. Assume \(a\le b\), and set
\[
D=b-a.
\]
For \(x\ge 0\), write \(x=q_xR+r_x\), with
\(q_x\in\mathbb{Z}_{\ge 0}\) and \(0\le r_x<R\), and define
\[
\Pi_R(x)=q_xR^2+r_x^2,
\qquad
\nu_R(x)=
\begin{cases}
0,&x=0,\\
\left\lceil x/R\right\rceil,&x>0.
\end{cases}
\]
Let \(j\) denote the total negative movement, so the total positive movement
is \(D+j\). Feasibility requires
\[
0\le j\le a,
\qquad
0\le j\le RS-b,
\]
and the positive and negative movements must fit on disjoint samples:
\[
\nu_R(D+j)+\nu_R(j)\le S.
\]
For fixed \(j\), the sum of squared differences is maximized by packing the
positive and negative movements into as few samples as possible, with each
sample carrying movement of magnitude at most \(R\). Therefore
\[
M_{\max}^{(R)}(a,b;S)
=
\max_{j\in\mathcal J(a,b)}
\left\{
\Pi_R(D+j)+\Pi_R(j)
\right\},
\]
where
\[
\mathcal J(a,b)
=
\left\{
j\in\mathbb{Z}_{\ge 0}:
j\le a,\;
j\le RS-b,\;
\nu_R(D+j)+\nu_R(j)\le S
\right\}.
\]
The objective is nondecreasing in \(j\), so the maximum is attained at the
largest feasible \(j\). The case \(a>b\) follows by symmetry.

For binary success/failure outcomes, \(R=1\), this reduces to
\[
q_{\max}^{(1)}(a,b;S)
=
\min(a+b,\;2S-a-b)-\frac{(b-a)^2}{S}.
\]
Here \(\min(a+b,2S-a-b)\) is the largest possible number of discordant samples
given the two binary task totals.

We allocate the marginal score totals \(A\) and \(B=A+L\) across tasks by
dynamic programming. Let \(F_j(u,v)\) be the largest attainable value of
\(Q\) after \(j\) tasks with \(u\) total score for policy \(A\) and \(v\)
total score for policy \(B\). Initialize
\[
F_0(0,0)=0,
\qquad
F_0(u,v)=-\infty
\quad\text{for }(u,v)\neq(0,0).
\]
For \(j=0,\ldots,T-1\), update
\[
F_{j+1}(u+a',v+b')
=
\max
\left\{
F_{j+1}(u+a',v+b'),
\;
F_j(u,v)+q_{\max}^{(R)}(a',b';S)
\right\}
\]
over every finite state \(F_j(u,v)\) and every
\[
0\le a'\le RS,
\qquad
0\le b'\le RS.
\]
Define
\[
Q_{\mathrm{hi}}(A,L;T,S,R)
=
F_T(A,A+L).
\]
Because the dynamic program enumerates every allocation of the two marginal
top-line counts across tasks, and because \(q_{\max}^{(R)}\) is the exact
task-level maximum for each allocation, \(Q_{\mathrm{hi}}\) is the exact
maximum of \(Q\) over all paired outcome tables compatible with top-line
counts \(A\) and \(B=A+L\). Thus every compatible paired outcome table
satisfies
\[
Q
\le
Q_{\mathrm{hi}}(A,L;T,S,R),
\]
and this upper bound is attained.

Therefore rejection is guaranteed whenever
\[
L
>
c_\alpha
\sqrt{
Q_{\mathrm{hi}}(A,L;T,S,R)
}.
\]
Conversely, if
\[
L
\le
c_\alpha
\sqrt{
Q_{\mathrm{hi}}(A,L;T,S,R)
},
\]
then the sharpness of \(Q_{\mathrm{hi}}\) gives a compatible paired outcome
table that does not reject.

The count-scale rejection-guarantee cutoff used for scalar summaries is the
suffix cutoff
\[
L_{\forall}(A;T,S,R,\alpha)
=
\min
\left\{
\ell\in\mathcal L_A:
\begin{array}{l}
\text{for every }L\in\mathcal L_A\text{ with }L\ge \ell,\\[2pt]
L
>
c_\alpha
\sqrt{
Q_{\mathrm{hi}}(A,L;T,S,R)
}
\end{array}
\right\},
\]
with the convention that the minimum is \(+\infty\) if the set is empty. The
normalized cutoff is
\[
\delta_{\forall}(\widehat\mu_A;T,S,R,\alpha)
=
\frac{L_{\forall}(A;T,S,R,\alpha)}{N}.
\]
Thus
\[
g
\ge
\delta_{\forall}(\widehat\mu_A;T,S,R,\alpha)
\]
implies that every paired outcome table compatible with the reported
top-line metrics rejects under the paired bounded-score Wald rule. For a
particular observed gap \(L/N\), the sharper pointwise guarantee condition is
\[
L
>
c_\alpha
\sqrt{
Q_{\mathrm{hi}}(A,L;T,S,R)
}.
\]

\subsection{Sharp Paired Classification}

For exact paired evaluations, the two envelopes above give a sharp
classification for each fixed count pair \((A,B=A+L)\):
\[
\begin{array}{ll}
L\le c_\alpha\sqrt{Q_{\mathrm{lo}}(L;T,S)}
&
\Longleftrightarrow
\text{no compatible paired outcome table rejects},
\\[6pt]
L>c_\alpha\sqrt{Q_{\mathrm{hi}}(A,L;T,S,R)}
&
\Longleftrightarrow
\text{every compatible paired outcome table rejects},
\\[6pt]
c_\alpha\sqrt{Q_{\mathrm{lo}}(L;T,S)}
<
L
\le
c_\alpha\sqrt{Q_{\mathrm{hi}}(A,L;T,S,R)}
&
\Longleftrightarrow
\begin{array}{l}
\text{some compatible paired outcome tables reject,}\\
\text{and some compatible paired outcome tables do not reject.}
\end{array}
\end{array}
\]
The middle line is the sharp top-line inconclusive case for paired
evaluations.

\subsection{Independently Sampled Evaluations}

Some published comparisons do not use the same paired benchmark instances for
both policies. Instead, each method is evaluated on independently sampled
task-conditioned instances, with the same \(T\) tasks and the same number
\(S\) of samples per task. In this setting, the paired difference statistic is
not the appropriate reference procedure. The corresponding stratified
two-sample Wald statistic uses separate within-task variance estimates for
the two policies.

Let
\[
a_t=\sum_{i=1}^S Y_{A,t,i},
\qquad
b_t=\sum_{i=1}^S Y_{B,t,i},
\]
and
\[
u_t=\sum_{i=1}^S Y_{A,t,i}^2,
\qquad
v_t=\sum_{i=1}^S Y_{B,t,i}^2.
\]
The two-sample denominator is determined by
\[
Q_{\mathrm{ind}}
=
\sum_{t=1}^T
\left[
u_t-\frac{a_t^2}{S}
+
v_t-\frac{b_t^2}{S}
\right].
\]
The count-scale stratified two-sample Wald statistic is
\[
Z_{\mathrm{ind}}
=
\frac{L}
{
\sqrt{
\frac{S}{S-1}Q_{\mathrm{ind}}
}
},
\]
with the same zero-variance convention as above.

For binary success/failure outcomes, this simplifies to
\[
Q_{\mathrm{ind}}
=
\sum_{t=1}^T
\left[
a_t-\frac{a_t^2}{S}
+
b_t-\frac{b_t^2}{S}
\right].
\]

The paired envelopes are conservative for this independently sampled setting.
For every independently sampled outcome table with top-line counts \(A\) and
\(B=A+L\),
\[
Q_{\mathrm{lo}}(L;T,S)
\le
Q_{\mathrm{ind}}
\le
Q_{\mathrm{hi}}(A,L;T,S,R).
\]

For the lower inequality, fix a task \(t\). Let \(X\) and \(Y\) be independent
draws from the empirical score distributions of policies \(A\) and \(B\) on
that task. Then \(Y-X\) is integer-valued,
\[
\mathbb{E}[Y-X]=\frac{b_t-a_t}{S},
\]
and
\[
S\operatorname{Var}(Y-X)
=
u_t-\frac{a_t^2}{S}
+
v_t-\frac{b_t^2}{S}.
\]
Among integer-valued random variables with a fixed mean, the variance is
minimized by placing mass on the two adjacent integers around the mean.
Therefore, with \(\rho_t=\rho_S(b_t-a_t)\),
\[
u_t-\frac{a_t^2}{S}
+
v_t-\frac{b_t^2}{S}
\ge
\rho_t-\frac{\rho_t^2}{S}.
\]
Combining residuals across tasks as in the paired lower-envelope proof gives
\[
Q_{\mathrm{ind}}\ge Q_{\mathrm{lo}}(L;T,S).
\]

For the upper inequality, again fix a task \(t\). Center the two empirical
score lists by subtracting their task means. Over a uniformly random pairing
of the \(A\)-scores and \(B\)-scores within the task, the sample covariance
has expectation zero. Hence there exists at least one pairing with covariance
\(C_t\le 0\). For that pairing, the paired task contribution satisfies
\[
Q_{\mathrm{pair},t}
=
\left(
u_t-\frac{a_t^2}{S}
+
v_t-\frac{b_t^2}{S}
\right)
-2C_t
\ge
u_t-\frac{a_t^2}{S}
+
v_t-\frac{b_t^2}{S}.
\]
Applying this argument independently within each task gives a compatible
paired outcome table whose \(Q\) is at least \(Q_{\mathrm{ind}}\). Since
\(Q_{\mathrm{hi}}(A,L;T,S,R)\) is the maximum paired value, it follows that
\[
Q_{\mathrm{ind}}\le Q_{\mathrm{hi}}(A,L;T,S,R).
\]

Thus the paired cutoffs remain valid as conservative top-line cutoffs for
independently sampled evaluations:
\[
L
\le
c_\alpha\sqrt{Q_{\mathrm{lo}}(L;T,S)}
\quad
\Longrightarrow
\quad
\text{no compatible independently sampled outcome table rejects},
\]
and
\[
L
>
c_\alpha\sqrt{Q_{\mathrm{hi}}(A,L;T,S,R)}
\quad
\Longrightarrow
\quad
\text{every compatible independently sampled outcome table rejects}.
\]
However, the intermediate region is not sharp for independently sampled
evaluations. It means only that the paired top-line bounds do not resolve the
corresponding stratified two-sample Wald test. Sharper two-sample top-line
bounds could classify some cases inside this region.

\subsection{Non-Realizable Metric Rounding}

The derivations above assume that the reported metrics determine exact integer score totals. Some methods, for example ones with non-deterministic policies, may report an aggregate score averaged over multiple runs over the test set with different seeds. Such reported scores may not be realizable from a single run over the test set. Address this by applying a convention that rounds the reported score to the nearest score realizable with a standard pass over the test set, and then treating it as we would the standard case. 

\section{Citation Tracker Methodology}
\label{app:tracker-methodology}

We use a citation tracker repository to derive both the count of new arXiv papers reporting numbers on a given benchmark in a given month (the ``79 new arXiv papers'' figure in the introduction) and the per-benchmark cumulative counts in Figure~\ref{fig:benchmark-result-paper-counts}.
Per-paper evidence and the full row-level audit are available through the project website: \href{https://ripl.github.io/manipulation_benchmark_audit/}{ripl.github.io/manipulation\_benchmark\_audit}.
The snapshot we report on was taken on 2026-05-21.

\subsection{Sources}
Candidates were harvested from Semantic Scholar (citation graph and term search), OpenAlex (citation graph, term search, DOI metadata), CORE (term and full-text search), the arXiv API (term search and metadata), OpenReview, Europe PMC, OpenCitations / COCI (citation-graph spot checks), OpenAIRE / ScholeXplorer (DOI / arXiv linkages), Crossref (DOI metadata and references), DataCite (DOI metadata, including arXiv DOIs), DBLP, NeurIPS proceedings pages, and project, code repository, and documentation pages.
We did not use Google Scholar / Publish or Perish exports, Lens.org, Scopus, Web of Science, or Dimensions.
Any paper indexed only by an excluded source is missed; the reported counts are therefore lower bounds.

\subsection{Deduplication}
Candidates are deduplicated by DOI, arXiv ID, Semantic Scholar paper ID, OpenAlex work ID, and normalized title.
A candidate surfaced by multiple sources is kept once; all source-level matches are preserved in a \texttt{Found via} column for row-level auditability.

\subsection{Per-Paper Classification}
Every candidate carries a \emph{usage} classification per benchmark:
\begin{list}{$\bullet$}{%
\setlength{\leftmargin}{1.5em}
\setlength{\itemsep}{0pt}
\setlength{\parsep}{0pt}
\setlength{\topsep}{2pt}
\setlength{\partopsep}{0pt}
}
\item \texttt{reports-results}: the paper reports at least one numeric success-rate (or task-completion) value for the benchmark, with the specific protocol or suite identified. This requires row-level full-text audit, not just a citation edge or a term match.
\item \texttt{citation-only}: the paper cites the benchmark seed paper but reports no numeric results.
\item \texttt{term-only-not-counted}: the paper matches a term search for the benchmark name but lacks a confirmed citation edge or reported result. Retained for source-coverage transparency; not counted.
\item \texttt{uses-data-no-results}: the paper trains on or otherwise reuses benchmark data without running its standard evaluation.
\item \texttt{unclear/full-text-unavailable}: full text could not be retrieved at audit time.
\item \texttt{seed-work}: the benchmark seed paper itself.
\end{list}
Only \texttt{reports-results} rows are counted as ``papers that reported numbers on the benchmark.''
The promotion rule is one-directional: we do not infer reported results from a citation or a name mention alone, and collision-prone names (``LIBERO,'' ``CALVIN,'' ``Habitat,'' and similar short tokens) stay at \texttt{term-only-not-counted} until row-level evidence promotes them.

\subsection{Date Convention}
Monthly bucketing uses the first arXiv posting date (v1), not the venue publication date.
We chose this because the venue date can lag the public claim by months to years, and benchmark adoption tracks the public claim.
Rows without an arXiv ID are excluded from monthly bucketing.

\subsection{Reproducing the Headline Count}
The introduction's ``79 new arXiv papers reported numbers on LIBERO in March 2026'' is the number of rows in the LIBERO tracker for which \texttt{First arXiv date} begins with \texttt{2026-03} and \texttt{LIBERO usage} equals \texttt{reports-results}.
At the 2026-05-21 snapshot, the LIBERO tracker contained 1{,}053 candidate rows, of which 551 are classified \texttt{reports-results}.
March 2026 contributes 79 of those, drawn from 117 March 2026 candidates overall: the remaining 38 are predominantly \texttt{citation-only}, with a smaller number in the other classes.

\subsection{Figure~\ref{fig:benchmark-result-paper-counts}}
Figure~\ref{fig:benchmark-result-paper-counts} applies the same filter to CALVIN, LIBERO, SimplerEnv, RoboCasa, and RoboTwin~2.0, bucketed by first arXiv date from July~2023 through March~2026.
Each benchmark uses an independent tracker, populated by the same source pipeline and classified under an analogous \texttt{$\langle$benchmark$\rangle$ usage} column.
A paper that reports results on multiple benchmarks is counted once per benchmark.

\end{document}